\pdfoutput=1

\documentclass[11pt]{article}

\usepackage[]{ACL2023}

\usepackage{times}
\usepackage{latexsym}

\newcommand{\update}[1]{{#1}} 

\usepackage[T1]{fontenc}

\usepackage[utf8]{inputenc}

\usepackage{microtype}
\usepackage{inconsolata}
\usepackage{graphicx}
\usepackage{subcaption}
\usepackage{changes}
 \usepackage{comment}
\usepackage{amsmath}
\usepackage{booktabs}

\definechangesauthor[name={Caio}, color=purple]{caio}

\title{Small Character Models Match Large Word Models for Autocomplete Under Memory Constraints}

\author{Ganesh Jawahar$^{\clubsuit}$\thanks{\:\:Part of work was done as an intern in Microsoft.}, Subhabrata Mukherjee$^\spadesuit$, Debadeepta Dey$^\spadesuit$, \\ \textbf{Muhammad Abdul-Mageed$^{\clubsuit\diamondsuit}$, Laks V.S. Lakshmanan$^{\clubsuit}$, Caio Cesar Teodoro Mendes$^\spadesuit$,} \\ \textbf{Gustavo Henrique de Rosa$^\spadesuit$, Shital Shah$^\spadesuit$}\\ \normalsize $^\clubsuit$University of British Columbia, $^\spadesuit$Microsoft $^\diamondsuit$MBZUAI \\
  \texttt{\small ganeshjwhr@gmail.com,  \{laks,amuham01\}@cs.ubc.ca}, \\ \texttt{\small \{Subhabrata.Mukherjee,dedey,caiocesart,gderosa,shitals\}@microsoft.com}}

\begin{document}
\maketitle
\begin{abstract}
Autocomplete is a task where the user inputs a piece of text, termed \textit{prompt}, which is conditioned by the model to generate semantically coherent continuation. 
Existing works for this task have primarily focused on datasets (e.g., email, chat) with high frequency user prompt patterns (or \textit{focused prompts})  where word-based language models have been quite effective. 
In this work, we study the more challenging 
setting consisting of low frequency user prompt patterns (or \textit{broad prompts}, e.g., prompt about {\tt \small $93^{rd}$ academy awards}) and demonstrate the effectiveness of \textit{character-based} language models. 
We study this problem under memory-constrained settings (e.g., edge devices and smartphones), where character-based representation is effective in reducing the overall model size (in terms of parameters). 
We use WikiText-103 benchmark to simulate broad prompts and demonstrate that character models rival word models in exact match accuracy for the autocomplete task, when controlled for the model size. For instance, we show that a $20$M parameter character model performs similar to an $80$M parameter word model in the vanilla setting. We further propose novel methods to improve character models by incorporating inductive bias in the form of compositional information and representation transfer from large word models. {Datasets and code used in this work are available at \url{https://github.com/UBC-NLP/char_autocomplete}.} 
\end{abstract}

\section{Introduction}
\label{sec:intro}

Autocomplete models are conditioned on user-written prompts or text to generate semantically coherent continuations. 
For example, given the user input ``{\tt \small Filmmaker George Lucas used Tikal as a \rule{0.5cm}{0.1mm}}'', a semantically coherent continuation can be ``{\tt \small filming location}'' (Example 1). Autocomplete models can dramatically reduce keystrokes and improve user's productivity in a wide range of applications including email, chat and document authoring. Some typical challenges in building a real-time autocomplete model include: (i) processing arbitrary length user input (e.g., paragraphs), 
(ii) handling low frequency user prompt patterns (or \textit{broad prompts}
 \update{that typically cover wider vocabulary} (as in Example 1), and (iii) satisfying memory constraints of the target device (such as cap on peak memory utilization). 

Despite the importance of the task, there has been limited research on autocomplete. Existing works such as Smart Compose~\cite{smartcompose_kdd19} and 
~\cite{trajanovski-etal-2021-text} train autoregressive language models on 
emails and chats, where user prompt patterns tend to be high-frequency. That is, the prompts are  \textit{focused prompts}, e.g., {a prompt about \tt \small office standups}, \update{that typically cover narrower vocabulary}. All these models are trained at word level, which leads to two issues: 
(i) input/output embedding parameters (less compressible component of the Transformer model~\cite{shen_aaai20}\footnote{\newcite{shen_aaai20} study the effects of quantization on different components of Transformer model, on the performance in various NLP tasks. They find that the embedding layer is most sensitive to quantization than other components and requires more bits to keep performance loss acceptable.}) occupy a significant share (e.g., more than 77\%) of the parameter budget due to the large vocabulary size and (ii) tendency to memorize high-frequency prompt patterns resulting in poor generalization on the low-frequency ones. 

\begin{table}[htb]
\footnotesize
\begin{center}
\begin{tabular}{c|c|c|c} 
\toprule
n-gram & unigram & bigram & trigram \\ \midrule 
Wikitext-103 & 95.44 & 84.35 & 60.63 \\
Reddit & 86.41 & 77.04 & 54.36 \\ \bottomrule
\end{tabular}
\vspace{-0.6em}
\caption{Percentage of unique out of vocabulary (OOV) n-grams in test set of WikiText-103 (broad prompts) vs. Reddit (focused prompts) datasets.}
\vspace{-0.6em}
\label{tab:oov}
\end{center}
\end{table}

In this paper, we focus on the autocomplete task 
of broad prompts from domains such as Wikipedia, where user prompt patterns often have low frequency (e.g., {prompt about \tt \small $93^{rd}$ academy awards}). For instance, from Table~\ref{tab:oov}, we observe that WikiText-103 (broad prompts) contains at least $10\%$ more unique out of vocabulary (OOV) n-grams compared to the Reddit dataset (focused prompts). 
This makes our task more challenging than conventional settings considered in prior work which do one of the following: (i) adopt word-based models that are good at memorizing high-frequency patterns for \textit{focused prompts} or (ii) rely on \textit{conventional language modeling} which is not geared for generating precise and short horizon continuations (see Section~\ref{sec:charvsword}).  

Furthermore, we study this problem for practical applications under memory-constrained settings. 
Lower-end edge platforms (e.g., Raspberry Pi with 256MB of memory~\cite{tinytl}) have  memory constraints that are more limiting than latency constraints, for supporting various on-device models. 
Also, given that autoregressive language models are memory-bounded~\cite{spatten_hpca21}, we focus on improving the accuracy-memory trade-off for autocomplete task of broad prompts. Our work is complementary to existing works in model compression including those on pruning~\cite{gordon2020compressing}, quantization~\cite{HanMao16} and distillation~\cite{sanh2019} that primarily focus on natural language understanding tasks (e.g., text classification). In contrast to these works, we study the effectiveness of character-based language models for a natural language generation task (e.g., autocomplete). 

\update{In this paper, we focus on two research questions. \textbf{RQ1}: How do character-based autocomplete models compare against word counterparts under memory constraints? \textbf{RQ2}: How to improve character-based autocomplete models with no negative impact on memory? We answer \textbf{RQ1} by showing that} compared to word models, character models (i) contribute $96\%$ fewer parameters in the embedding layer due to a much smaller vocabulary, (ii) work well on low-frequency (or broad) prompt patterns (e.g., $21$\% accuracy improvement by using $20$M character model over $20$M word model, see Figure~\ref{fig:acc-mem-pareto} (a)) and (iii) result in high savings on peak memory utilization (e.g., $4.7$\% memory savings by  using $20$M character model over $20$M word model, see Figure~\ref{fig:acc-mem-pareto} (b)). 
When controlled for model size (number of parameters), we find that 
smaller character models (e.g., $20$M parameters) perform similar to large word models (e.g., $80$M parameters).  
\update{We answer \textbf{RQ2}} by developing novel methods to improve the accuracy of character models, which unlike previous work, have \textit{minimal impact on  memory usage}. These methods introduce inductive bias in the form of compositional information and representation transfer from large word models (best method). 
We show that the best method achieves $1.12$\% and $27.3$\% {relative} accuracy improvements over vanilla character and vanilla word models respectively with no impact on memory usage. We discuss the limitations of our work in Section~\ref{sec:limitation} and defer the analysis of accuracy-latency trade-off to future work while focusing only on memory-constrained settings in this work.

Our major contributions are as follows: \textbf{(1)} To the best of our knowledge, this is the first study of the  autocomplete task for broad prompts in a memory-constrained setting. \textbf{(2)} We perform an extensive comparison of character and word models across diverse architectures and demonstrate the advantage of character models over large word models for the autocomplete task on dimensions like peak memory utilization and model parameters. \textbf{(3)} We introduce novel methods leveraging inductive bias to further improve the accuracy of character models with minimal impact on  memory usage.

\section{Related Work}
\label{sec:related_work}
Our work leverages advances in neural language models, autocompletion, and efficient deep learning. 

\noindent\textbf{{Neural Language Models.}} The autocomplete models we study in this work utilize Transformer-based~\cite{vaswani_neurips17} autoregressive neural language models as backbone. 
Compared to word models, character models lag behind in language modeling performance when controlled for model size~\cite{AlRfou2019CharacterLevelLM,Choe2019BridgingTG} and have a high computational complexity due to long sequence length~\cite{tay2021charformer}. In this work, we focus on deploying models on lower-end edge platforms (e.g., Raspberry Pi) where memory, as opposed to latency, is the major bottleneck. 

\noindent\textbf{{Autocomplete Task.}} Despite the pervasiveness of autocomplete models, there is limited research in the academic community on the autocomplete task. 
Gmail Smart Compose~\cite{smartcompose_kdd19} is a popular word-based autocomplete model 
for email suggestions.  
They find the encoder-decoder architecture to have a higher latency than the decoder-only architecture. They also find the Transformer architecture to be marginally better than the LSTM architecture~\cite{lstm}. Motivated by these findings, we employ a decoder-only, Transformer based architecture for building our autocomplete model. \newcite{trajanovski-etal-2021-text} leverage word-based autocomplete models for providing email and chat suggestions. 


In this work, we focus on building autocomplete models for broad prompts from domains such as Wikipedia, where user prompt patterns can be quite low frequency (e.g., {prompt about  {\tt\small Bruce Vilanch} (Oscars writer), with frequency of only 6 times}). Unlike our prompt completion task, query autocompletion task is a well researched problem~\cite{qcomplete0,qcomplete3,qcomplete1,gog_effqa}, where the goal is to complete the user's query, e.g., search  query. Since user queries are generally short, query autocomplete models need not track long-range dependencies to understand the user's intent. In contrast, it is a \textit{requirement} in our prompt completion setting, as the user prompt can be arbitrarily large, e.g., sentences or paragraphs.

\update{ChatGPT~\cite{chatgpt} and GPT-4~\cite{gpt4} are recent dialogue models, which have garnered a great attention from the AI community for their ability to converse with human-like capabilities. The data used to train these models are not disclosed by the authors. As it is entirely possible for their training data to include the test sets we study in our work and train-test overlap analysis cannot be performed, we cannot make a fair comparison of our work with these `closed' AI models~\cite{rogers-etal-2023-closed}. Models such as Alpaca~\cite{alpaca}, Vicuna~\cite{vicuna2023}, GPT-4-LLM~\cite{gpt4llm} that claim to perform similarly as ChatGPT  with few billion parameters are usually finetuned with outputs from ChatGPT or GPT-4. Hence, these models cannot be fairly compared with our work either.}

\noindent\textbf{{Efficient Deep Learning.}} 
Exponential growth in the size of Transformer-based autoregressive language models (e.g., $175$B~\cite{gpt3})
has given rise to a strong need to make these 
models efficient so they can be used on commodity devices like laptop, tablet, and mobile, which have various resource constraints such as peak \textit{memory} utilization and \textit{latency}, while yielding the best performance under the  constraints. To this end, there has been extensive research on building efficient Transformer models that are smaller, faster, and better, 
as summarized thoroughly by~\newcite{tay_effsurvey} and~\newcite{menghani_effsurvey}. Our work is focused on improving the efficiency of a natural language generation task (e.g., autocomplete), which has received less attention from an efficiency perspective.~\newcite{spatten_hpca21} observe that 73\% of the overall latency of autoregressive language models goes to memory intensive data movement operations (e.g., splitting heads, transpose, reshape) and conclude that these models are memory intensive.
Since lower-end edge platforms have tighter memory constraints than latency constraints~\cite{tinytl}, \textit{we focus on improving the  accuracy-memory trade-off of autocomplete models}. 

\section{Autocomplete -- Fundamentals}
\label{sec:autocomp_problem}

\noindent\textbf{{Problem.}} Given a text sequence $\mathbf{x} = (x_1,\dots,x_{|\mathbf{x}|})$ (user input) with tokens from a fixed vocabulary $x_i \in \mathcal{V}$, the goal of the autocomplete task is to generate a completion $\hat{\mathbf{x}}_{k+1:N}$ such that the resulting sequence ($x_1,\dots,x_k,\hat{x}_{k+1},\dots,\hat{x}_N$) resembles a sample from $p_*$, where $p_{*}(\mathbf{x})$ denotes the reference distribution. $\mathbf{x}$ can be arbitrarily large (e.g., paragraphs), while  $\hat{\mathbf{x}}_{k+1:N}$ is generally short (e.g., three words). Each token $x_k$ can be a  word, character, or subword. 
The vocabulary $\mathcal{V}$ contains unique tokens from the dataset $\mathcal{D}$ consisting of a finite set of text sequences from $p_*$.

\noindent\textbf{{Data.}} Most datasets in the autocomplete literature come from domains with focused prompts (e.g., emails~\cite{smartcompose_kdd19,trajanovski-etal-2021-text}, chat messages~\cite{trajanovski-etal-2021-text}). 
In this work, we target the autocomplete task on datasets 
with broad prompts (e.g., Wikipedia) with a lot of low-frequency prompt patterns (e.g., {the prompt  \tt \small EACL 2023 conference}). Autocomplete models trained to answer broad prompts can be used to assist users in completing documents such as essay, report, letter, etc. 

\noindent\textbf{{Metrics.}} The commonly used metric for evaluating the quality of an autocomplete model is ExactMatch@N~\cite{rajpurkar-etal-2016-squad} which measures the percentage of the first $N$ words in the predicted suggestion that exactly match  the first $N$ words in the ground truth suggestion. ExactMatch@Overall~\cite{smartcompose_kdd19} is a  weighted average of the ExactMatch for all subsequence lengths up to $K$. For our setting, larger n-grams are increasingly difficult to predict for both word and character models as shown in Figure~\ref{fig:n-vs-acc}. Hence   we set $K$ to 3. Since the exact match metric strictly looks for full match of the subsequence, it is a hard metric to improve on, especially for broad prompts. One can utilize a less stringent metric such as PartialMatch~\cite{trajanovski-etal-2021-text}. PartialMatch  measures the percentage of characters in the first $N$ words in the predicted suggestion that exactly match those of the ground truth suggestion. However, PartialMatch might not adequately penalize for the grammatical incorrectness of the predicted suggestion. 
\newcite{trajanovski-etal-2021-text} also utilize metrics 
that require interactions from real users, which are difficult to acquire in practice. Given that  the user-based metrics and PartialMatch metric have a strong correlation with ExactMatch in all the experiments carried out by~\newcite{trajanovski-etal-2021-text}, we use the exact match metric to quantify the performance of the autocomplete model in this work. We further perform human evaluation to compare the naturalness and user acceptability of the suggestions generated by different models.\footnote{For our final comparison, however, we report PartialMatch vs. ExactMatch (Table~\ref{tab:final_res}). We do not experiment with ranking metrics (e.g., mean reciprocal rank) since our autocomplete model produces just a single suggestion.}


\noindent\textbf{{Model.}} 
We adopt the Transformer architecture, specifically Transformer-XL~\cite{dai-etal-2019-transformer}, for our autocomplete model. 
 We choose Transformer-XL for the following two reasons: (i) as~\newcite{dai-etal-2019-transformer} show, the model achieves strong results on word and character-based language modeling benchmarks and (ii) the model can handle long text sequences (e.g., 1600 word tokens or 3800 character tokens) which is crucial for treating arbitrarily long user inputs ($\mathbf{x}$).

\noindent\textbf{{Training.}} 
We train a decoder-only, Transformer-XL model that conditions on user input to generate the suggestion autoregressively. 
The parameters $\theta$ of the autocomplete model $p_{\theta}(\mathbf{x})$ can be optimized using the standard language modeling objective.

\noindent\textbf{{Inference.}} During inference, the model $p_{\theta}(\mathbf{x})$ takes the user input $\mathbf{x}_{1:k} \sim p_*$ and generates the suggestion $\hat{\mathbf{x}}_{k+1:N} \sim p_\theta(.|\mathbf{x}_{1:k})$ such that ($x_1,\dots,x_k,\hat{x}_{k+1},\dots,\hat{x}_N$) resembles a sample from $p_*$. 
In this work, we choose greedy search and select the token that receives the highest probability as the generated token; that is, ${\hat{x}}_t = \arg \max p_\theta(x_t|x_1,\dots,x_{t-1})$. 
As shown in Appendix~\ref{sec:beam_search_analysis} (see Figure~\ref{fig:beam_search_analysis}), beam search performs poorly on our task and the trends we see in the next section do not depend on the choice of the decoding algorithm.  
For simplicity, we assume the  autocomplete model generates exactly one suggestion $\hat{\mathbf{x}}_{k+1:N}$.
\section{Character vs. Word Model}
\label{sec:charvsword}

\begin{figure*}[t!]
    \centering
    \begin{subfigure}[t]{0.4\textwidth}
        \centering
        \includegraphics[height=1.25in, width=2.0in]{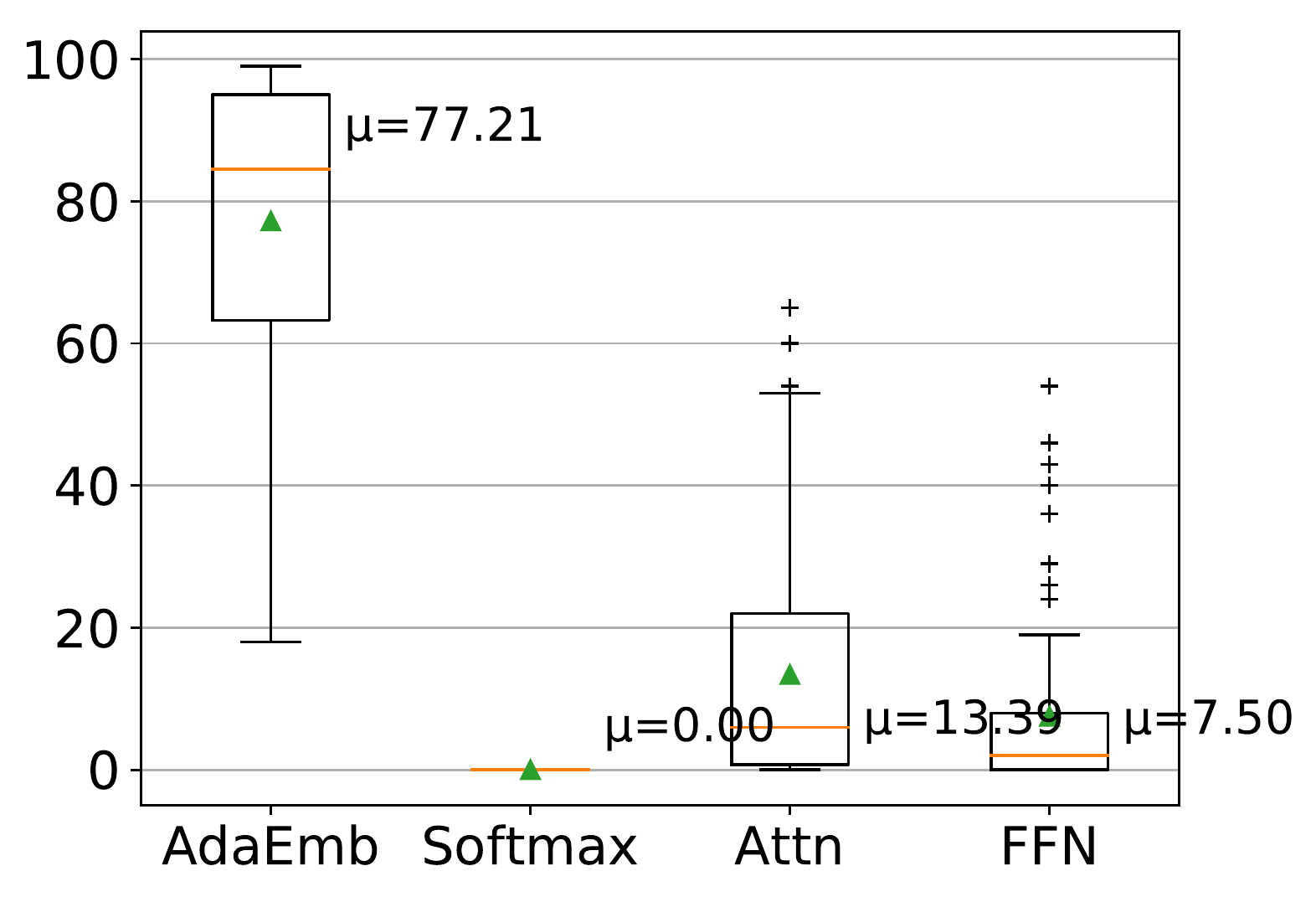}
        \caption{Word}
    \end{subfigure}%
    ~ 
    \begin{subfigure}[t]{0.4\textwidth}
        \centering
        \includegraphics[height=1.25in, width=2.0in]{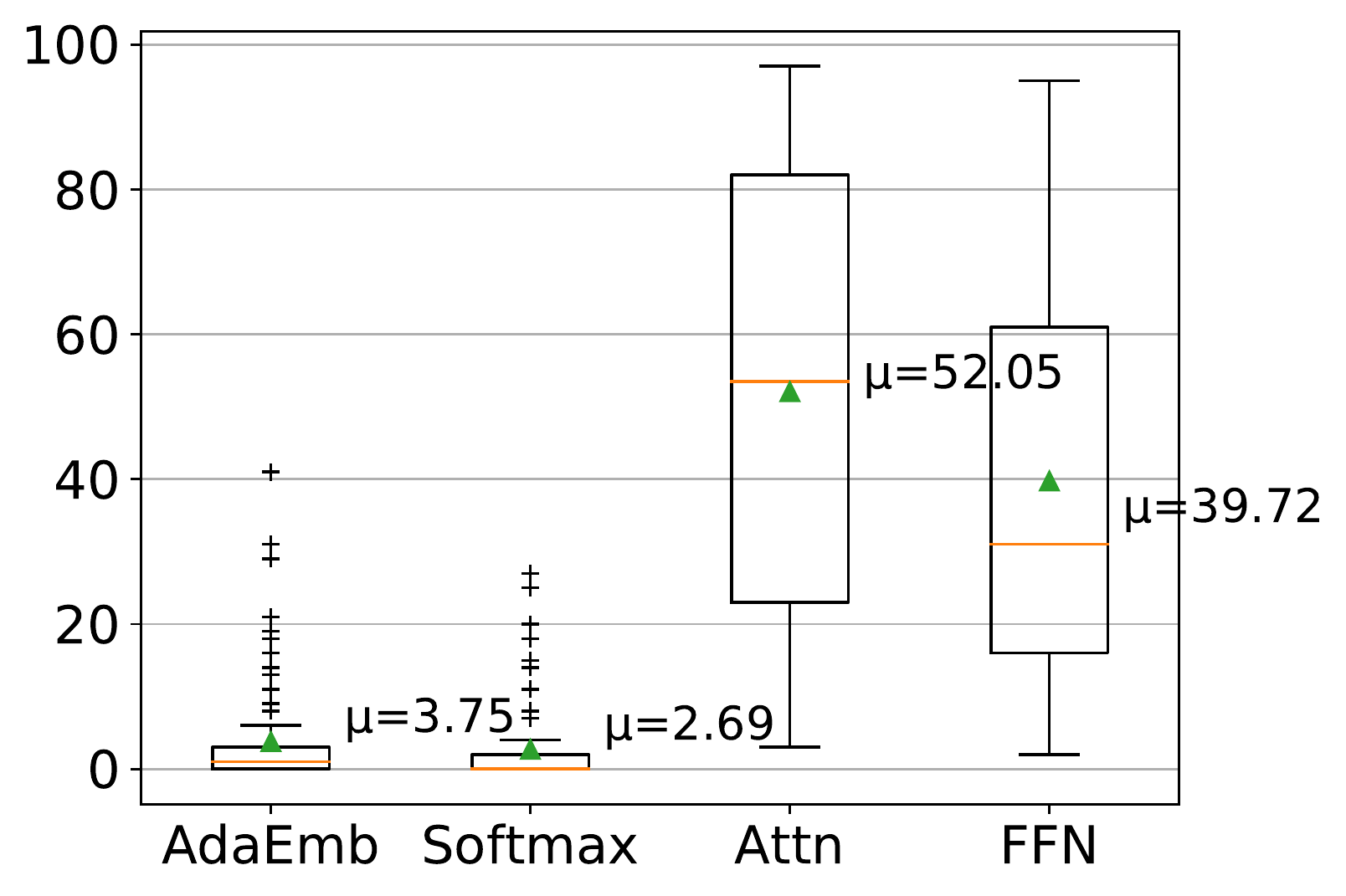}
        \caption{Character}
    \end{subfigure}
    \vspace{-0.5em}
    \caption{Percentage of parameters allocated to a given component w.r.t. different components in Transformer-XL model aggregated across $100$ random architectures.}
    \vspace{-0.5em}
    \label{fig:char-transxl-param-distrib}
\end{figure*}

Existing autocomplete models are primarily word-based, i.e., the representation choice for $x_k$ is word. Word-based autocomplete models have the following properties: (i) they invest most of the parameters (e.g., more than 77\%) 
from the overall parameter budget on the embedding layer, which is less likely compressible using standard techniques such as quantization~\cite{shen_aaai20} and (ii) they can memorize high-frequency prompt patterns and perform well on datasets with focused prompts (e.g., Reddit posts). 
\textit{In this work, we {focus on autocompletion on broad prompts} and we aim to keep the parameter allocation to the embedding layer as small as possible thereby improving the overall memory footprint.} 
To this end, we choose {\em character} as the representation choice and study the memory-accuracy tradeoff of character based models on the autocomplete task for broad prompts. Character-based autocomplete models have several desirable properties compared to their word based counterpart, as they (i) invest far fewer parameters (e.g., less than 4\%) of the parameter budget on the embedding layer and  invest most parameters on other highly compressible 
Transformer components such as self-attention network, feedforward network, and softmax layer; (ii) perform well on datasets with broad prompts (as we will show); and 
(iii) provide a better tradeoff between accuracy and memory (model size and peak memory utilization). To demonstrate these 
{properties,} we perform extensive experiments on the WikiText-103 benchmark~\cite{wikitext103} (unless stated otherwise). This benchmark contains about $100$M tokens from Wikipedia to simulate broad prompts. Since we focus on improving the memory footprint of autocomplete models, we do not experiment with subword models, which introduce a large number of token embeddings in the embedding layer (e.g., $50$K), compared to their character based counterpart. \update{In other words, we focus only on character models that keep the parameter allocation to the embedding layer as small as possible thereby improving the overall memory footprint.}

\begin{figure*}[t!]
    \centering
    \begin{subfigure}[t]{0.4\textwidth}
        \centering
        \includegraphics[height=1.25in, width=2.0in]{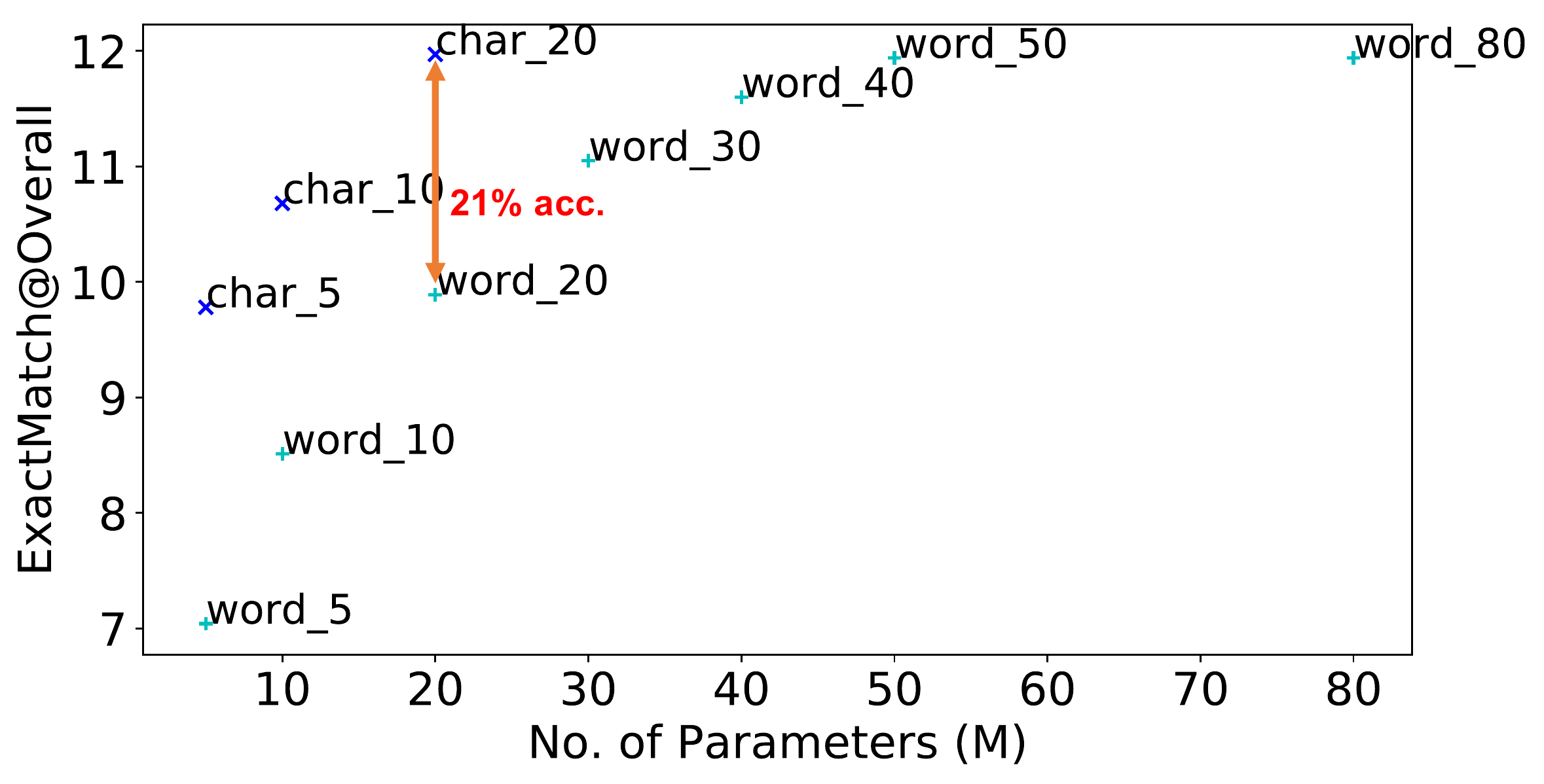}
        \caption{Accuracy vs. No. of Parameters}
    \end{subfigure}%
    ~ 
    \begin{subfigure}[t]{0.4\textwidth}
        \centering
        \includegraphics[height=1.25in, width=2.0in]{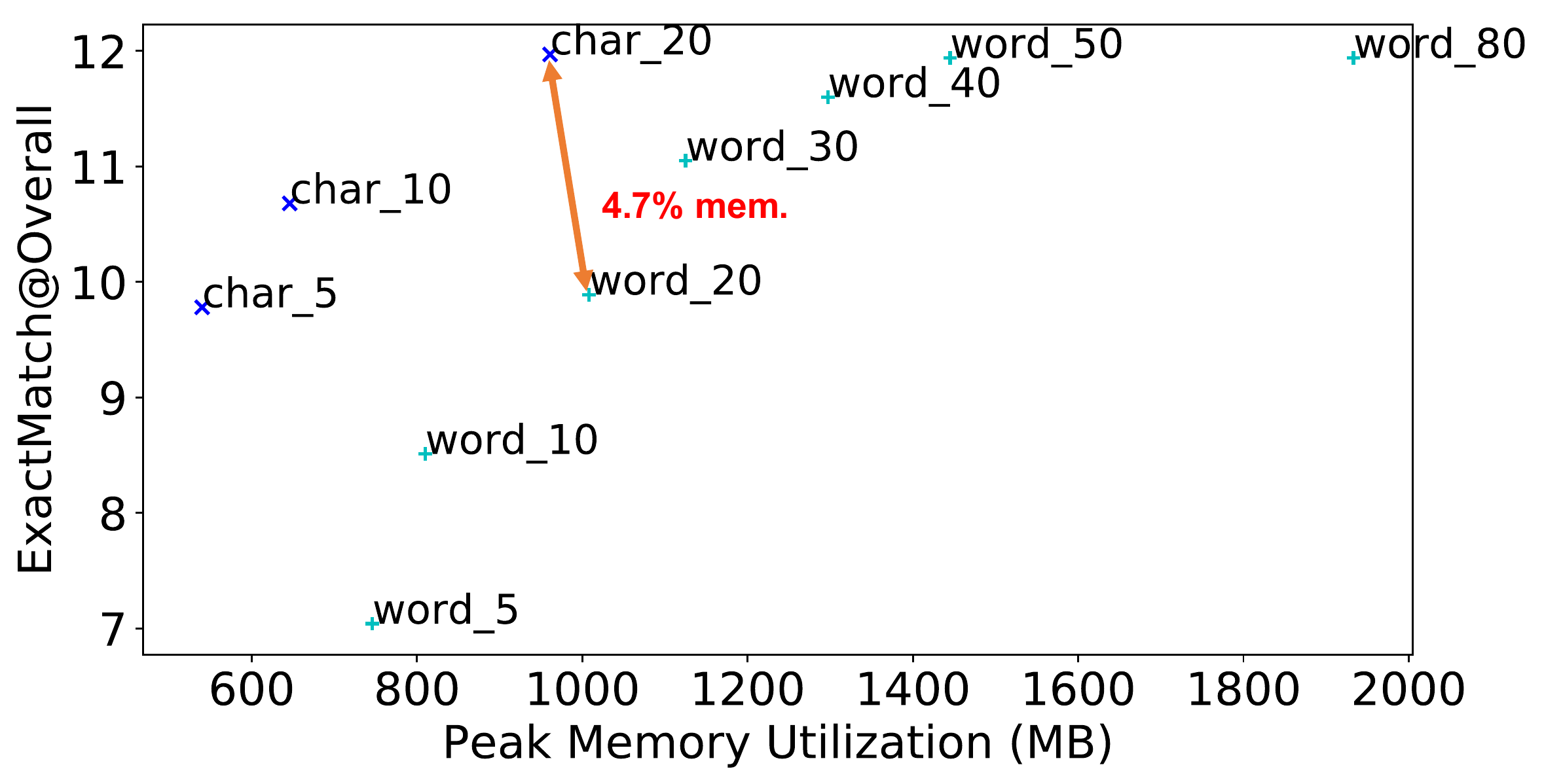}
        \caption{Accuracy vs. Peak Memory Utilization}
    \end{subfigure}
    \vspace{-0.5em}
    \caption{Accuracy-Memory Pareto Curve. Each point in the curve has number of model parameters at the end.}
    \vspace{-0.5em}
    \label{fig:acc-mem-pareto}
\end{figure*}

\begin{figure*}[t!]
    \centering
    \begin{subfigure}[t]{0.4\textwidth}
        \centering
        \includegraphics[height=1.25in, width=2.0in]{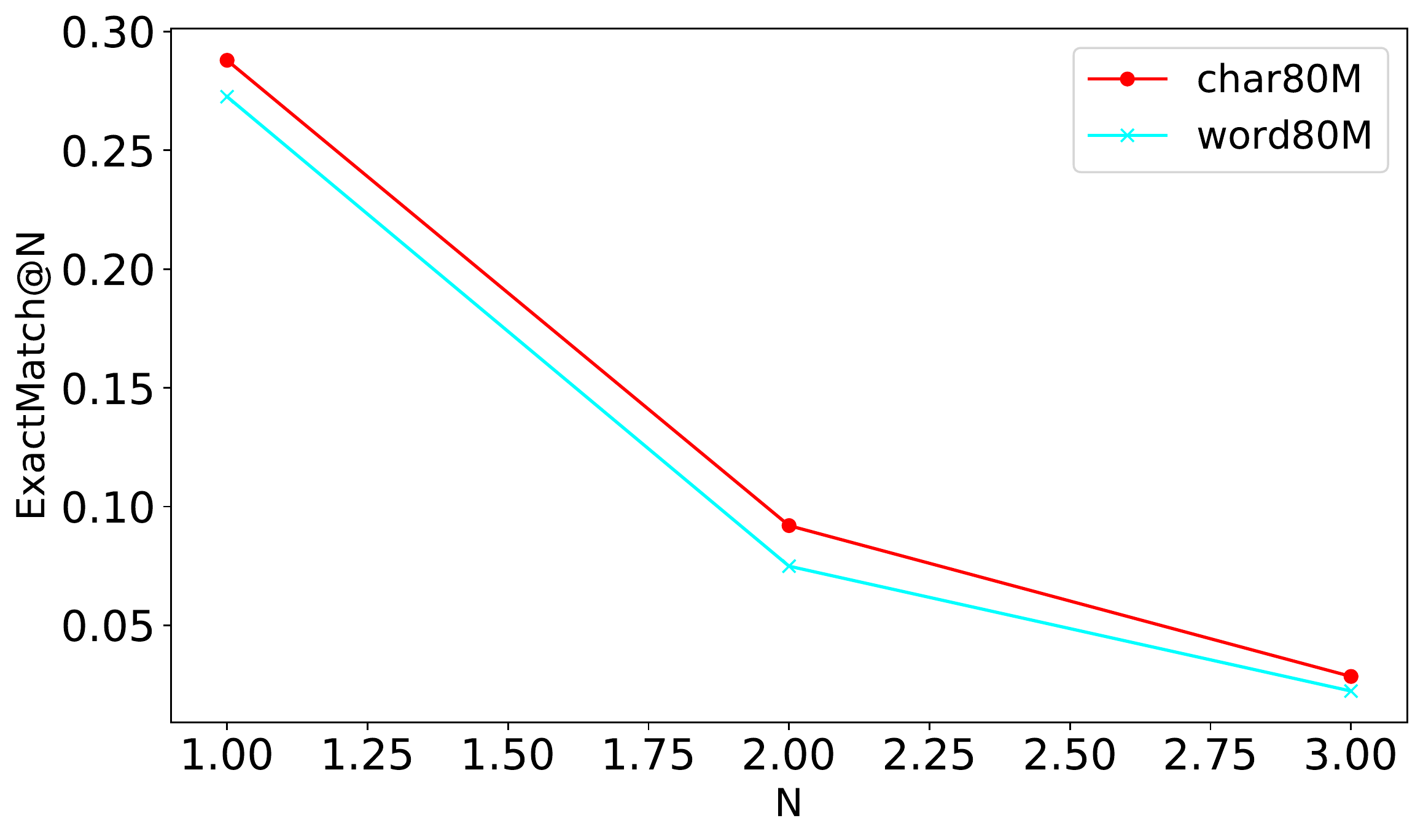}
        \caption{Wikitext-103}
    \end{subfigure}
    \begin{subfigure}[t]{0.4\textwidth}
        \centering
        \includegraphics[height=1.25in, width=2.0in]{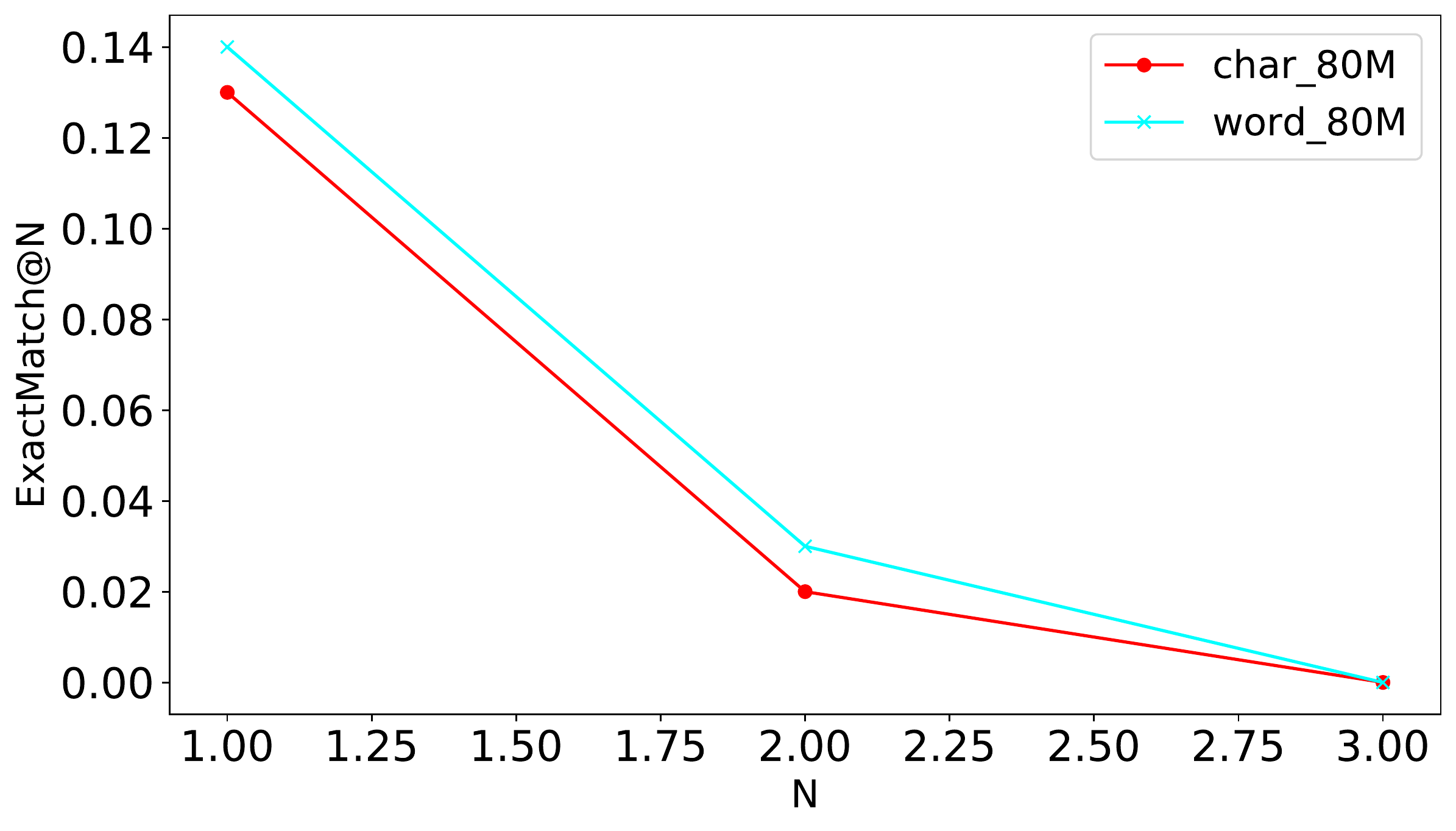}
        \caption{Reddit}
    \end{subfigure}
    \vspace{-0.5em}
    \caption{ExactMatch@N vs. N for word and char. model on first 500 samples from Wiki-103 and Reddit Dev sets.}
    \vspace{-0.5em}
    \label{fig:n-vs-acc}
\end{figure*}

\noindent\textbf{{Component-Wise Parameter Breakdown.}} Transformer-XL model can be broken down into four components: (i) adaptive embedding layers (AdaEmb)~\cite{baevski2018adaptive}, which contain shared input and output token embeddings; (ii) self-attention layers (Attn); (iii) feedforward network layers (FFN); and (iv) output softmax layers (Softmax). Figure~\ref{fig:char-transxl-param-distrib} shows the percentage of parameters allocated to each component for both word- and character-based models, averaged over 100 random architectures for each representation.\footnote{The hyperparameter space used to sample architectures is shown in Appendix~\ref{sec:hypspace_compo}.} Word-based models allocate more than $77\%$ 
of the parameters to the embedding layers, which are less amenable to compression, for purposes of generating efficient and smaller models. These models allocate less than $14\%$ and $8\%$ of the parameter budget to highly compressible layers such as self-attention and feedforward network layers. In contrast, character-based models allocate more than $90\%$ of the parameters to these highly compressible layers and less than $4\%$ to the embedding layers. Hence, character-based models have the potential to admit {much} greater compression using standard techniques such as distillation and quantization with a negligible performance drop.

\begin{figure}
\centering
\includegraphics[width=2.2in,height=1.25in]{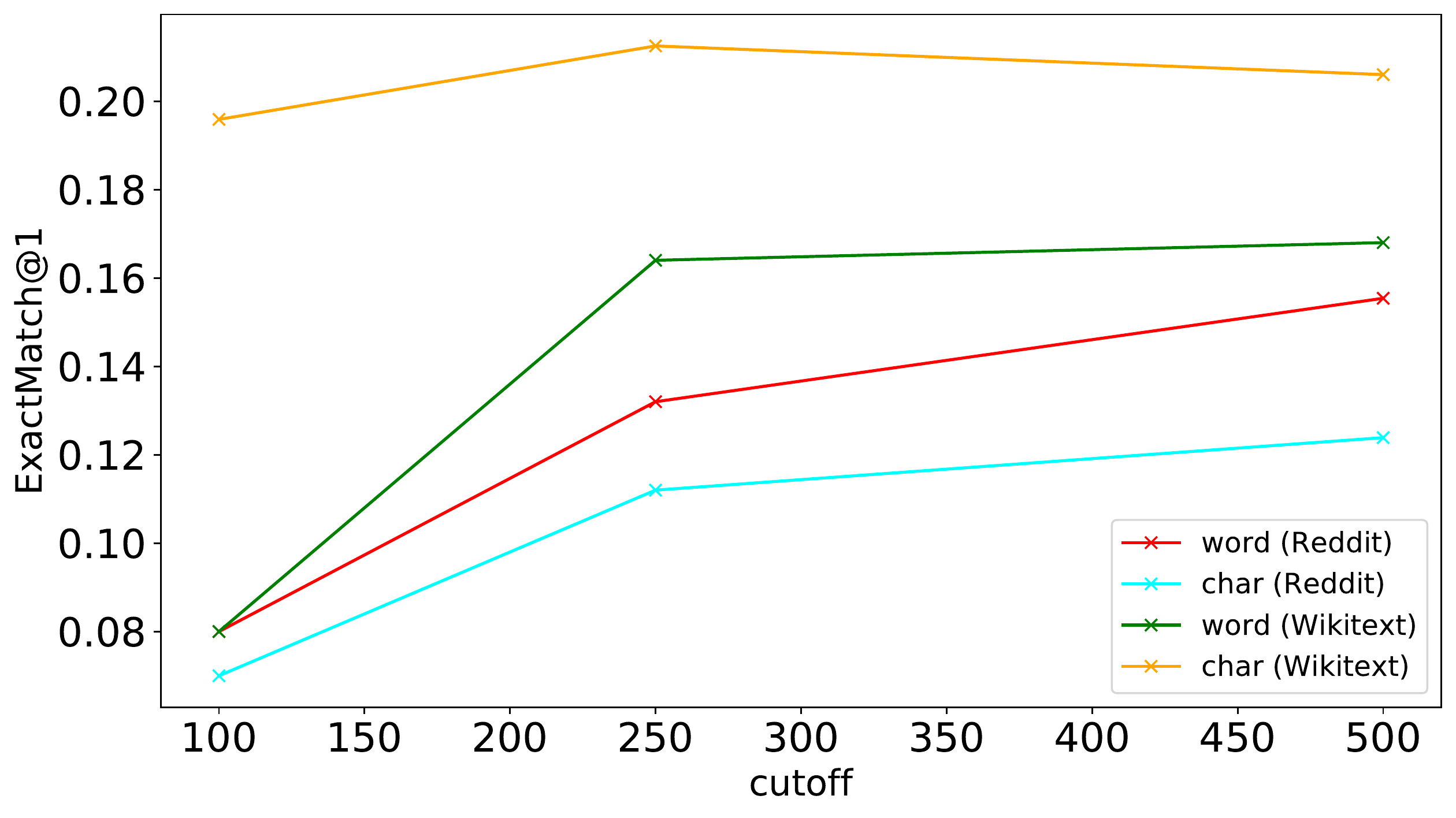}
\vspace{-0.5em}
\caption{ExactMatch@1 vs. Cutoff for word and character model. Cutoff refers to the top $k$ prompts based on the percentage of OOV n-grams (upto 3) in ascending (descending) order for WikiText (Reddit), where $k \in \{100, 250, 500\}$. Character models perform better than word models on WikiText (broad prompts) and vice versa on Reddit (focused prompts).}
\vspace{-0.5em}
\label{fig:binsize-accuracy}
\vspace{-0.5em}
\end{figure}

\noindent\textbf{{Accuracy vs. Memory Tradeoff.}}
Although character-based models seem to have better compression potential, their autocomplete performance gap over word-based models as a function of memory is not immediately obvious. We study the effect of memory in two ways: (i) model size, which corresponds to the total number of model parameters, and (ii) peak memory utilization, which measures the peak amount of memory utilized by a process during inference. 
In all our experiments, the decoding of character models stops once the desired number of words (identified by space character) are predicted.
\update{The hyperparameter values for word and character autocomplete models of different sizes can be seen in Table~\ref{tab:hypspace_word} and Table~\ref{tab:hypspace_char} respectively.}
Figure~\ref{fig:acc-mem-pareto} shows the accuracy-memory pareto curve\footnote{Hyperparameter values of different model sizes for word and character models can be found in Appendix~\ref{sec:hypspace_word_char}.}. 
Surprisingly, we observe that small character models (e.g., $20$M) can rival large word models (e.g., $80$M) in terms of accuracy-memory tradeoff. For instance, if we use a character model of size $20$M instead of a word model of size $80$M, we can save $75\%$ of the model parameters and more than $60\%$ of the peak memory utilization for a performance drop of $<0.5$ points. 

\begin{figure*}[t!]
    \centering
    \begin{subfigure}[t]{0.24\textwidth}
        \centering
        \includegraphics[height=0.7in, width=1.45in]{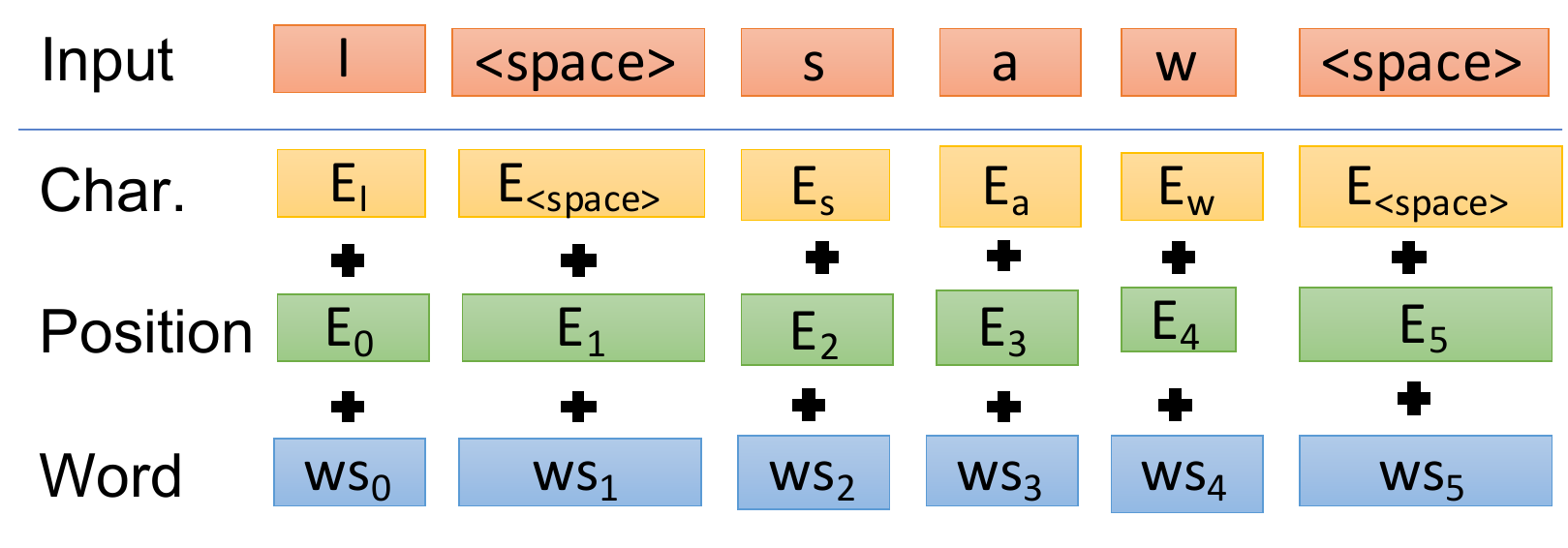}
        \caption{BERT-Style method}
    \end{subfigure}%
    ~ 
    \begin{subfigure}[t]{0.29\textwidth}
        \centering
        \includegraphics[height=0.7in, width=1.85in]{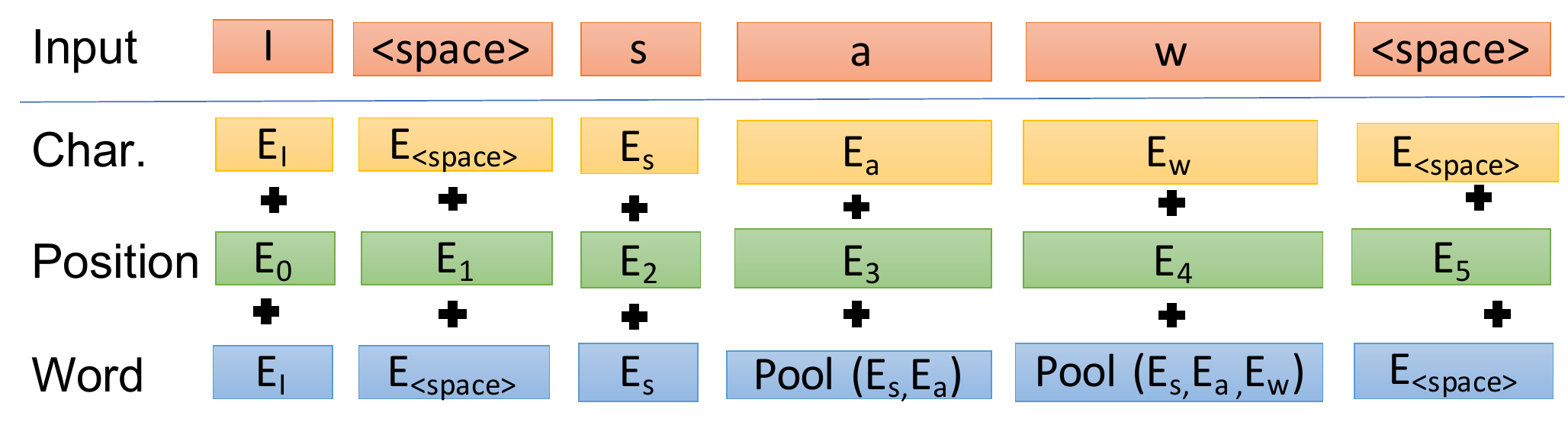}
        \caption{Character pooling method}
    \end{subfigure}
    ~
    \begin{subfigure}[t]{0.43\textwidth}
        \centering
        \includegraphics[height=0.7in, width=2.6in]{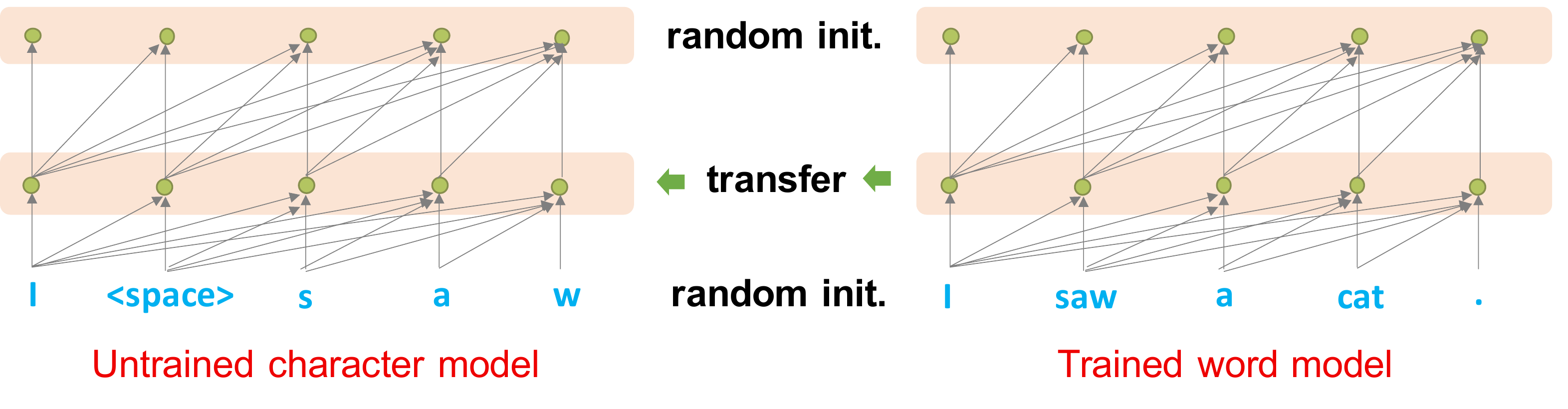}
        \caption{Transfer from word models method}
    \end{subfigure}
    \vspace{-0.5em}
    \caption{Methods to improve character models. Note `Position' in (a), (b) refers to character position embeddings.}
    \vspace{-0.5em}
    \label{fig:improv_methods}
\end{figure*}

\noindent\textbf{{Broad vs. Focused Domain.}} Prior works~\cite{AlRfou2019CharacterLevelLM,Choe2019BridgingTG} have found character models to be lagging behind word models in language modeling performance. Surprisingly,  small character models perform similarly to or better than big word models on the autocomplete task. We hypothesize that the reason behind the superior performance of character models in our setting is due to their ability to answer broad prompts 
better than word-based models. To validate this claim, we compare character and word models on their ability to answer broad and focused prompts, controlled for the model size consisting of $80$M parameters each.  

From Table~\ref{tab:oov}, we observe that the percentage of unique out-of-vocabulary (OOV) n-grams in WikiText-103 is $10\%$ higher than that in the Reddit dataset. While WikiText and Reddit by nature have a different vocabulary distribution, the significant gap in the relative proportions of OOV n-grams indicates that Wikipedia articles cover more diverse and broad domains. Therefore we simulate broad prompts using articles from WikiText-103 
and focused prompts with user posts from \url{Reddit.com} website {(The Pushshift Reddit Dataset~\cite{reddit_pushshift}, see Appendix~\ref{sec:reproduce} for more details).} As shown in Figure~\ref{fig:n-vs-acc}, the performance of the word-based model is superior to that of the character-based model in answering focused prompts, but not for answering broad prompts. A potential reason is the tendency of word-based models to memorize high-frequency patterns that are rife in datasets with focused prompts. On the other hand, character-based models excel on answering broad prompts (which are the focus of our work) which can be attributed to their superior ability in handling low-frequency patterns. We observe this trend with character-based models when we report the accuracy on the the top $k$ (`cutoff') low (high) frequent prompt patterns for WikiText (Reddit) selected by ranking the prompts based on the percentage of OOV n-grams (up to 3) in the ascending (descending) order (see Figure~\ref{fig:binsize-accuracy}). We also observe the trend for unseen datasets with broad prompts (e.g., Penn Treebank, see Appendix~\ref{sec:acc-vs-mem-unseen-datasets}). 

\section{Methods to Improve Character Models}
\label{sec:methods}

In the previous section, we demonstrated character-based models to be more efficient than word-based models for the autocomplete task on broad prompts. Unlike word-based models, which directly consume words, character-based models are forced to learn and compose semantically meaningful textual  
units (e.g., suffixes, words) from more granular lexical units in the form of characters. Therefore, methods that can explicitly integrate information from semantic units higher than characters (such as from words or word segments) can propel the performance of character based models~\cite{hoon_sigir17}. However, \textit{existing methods primarily focus on improving the accuracy of character models, often at the expense of memory}. For example, \newcite{hoon_sigir17} augment a character model with explicit model parameters for word embeddings, which add several millions of additional parameters (e.g., $13M$ parameters with modest embedding size of $50$ and standard WikiText-103 word vocabulary size of $267K$).  
We introduce some novel methods that explicitly integrate word information into the character model with negligible impact on memory, as discussed next.

\noindent\textbf{{BERT-Style Word Segment Embedding.}} 
In this method, we introduce a word segment embedding layer which acts as an inductive bias by providing the word segment information explicitly in addition to character and position embedding layers (Figure~\ref{fig:improv_methods} (a)). This word segment embedding layer is inspired by the sentence segment layer of BERT~\cite{devlin-etal-2019-bert} which helps the model distinguish sentences in the textual input. In our case, the word segment embedding layer can help the model distinguish words in the textual input. The number of additional model parameters introduced by this layer {equals} the maximum number of words in a training input sequence times the embedding dimension, which is generally negligible. 

\noindent\textbf{{Character Pooling.}} 
In this method, we compute word embeddings by pooling from embeddings of characters seen so far for the current word (see Figure~\ref{fig:improv_methods} (b)). The pooling function takes a set of character embeddings as input, and outputs the word embedding which is concatenated with other embeddings (as additional input) similar to the previous method. We experiment with non-parameterized, simple pooling functions such as sum, mean, and maximum. 
Unlike the previous method, the character pooling method does not introduce additional model parameters, due to the choice of our pooling function. The computation of word embedding does not involve look-ahead embeddings from characters belonging to the current word (that are not seen at the current timestep), thus preventing data leakage that could render the language modeling task trivial.

\noindent\textbf{Transfer from Word Models.} In this method, we initialize a subset of decoder layers of the character model with decoder layers from a trained word model. Unlike previous methods, the decoder layer transfer method can appropriately exploit the rich syntactic and semantic information learned by the word model, which serves as a good starting point for training a character model rather than training from scratch. Figure~\ref{fig:improv_methods} (c) illustrates the transfer of the bottom $50\%$ of decoder layers from the word model to the character model. Similar to the character pooling method, this method does not introduce additional model parameters. Rather, this method introduces a novel hyperparameter that controls the percentage of word-level bottom layers to transfer into our character-level model, which is tuned on the validation set. To the best of our knowledge,   no prior work  has explored transferring layers from a source trained model, where the source and the target model have very different vocabularies.


\section{Results}
\label{sec:results}

\begin{figure}
\centering
\includegraphics[width=2.3in,height=1.25in]{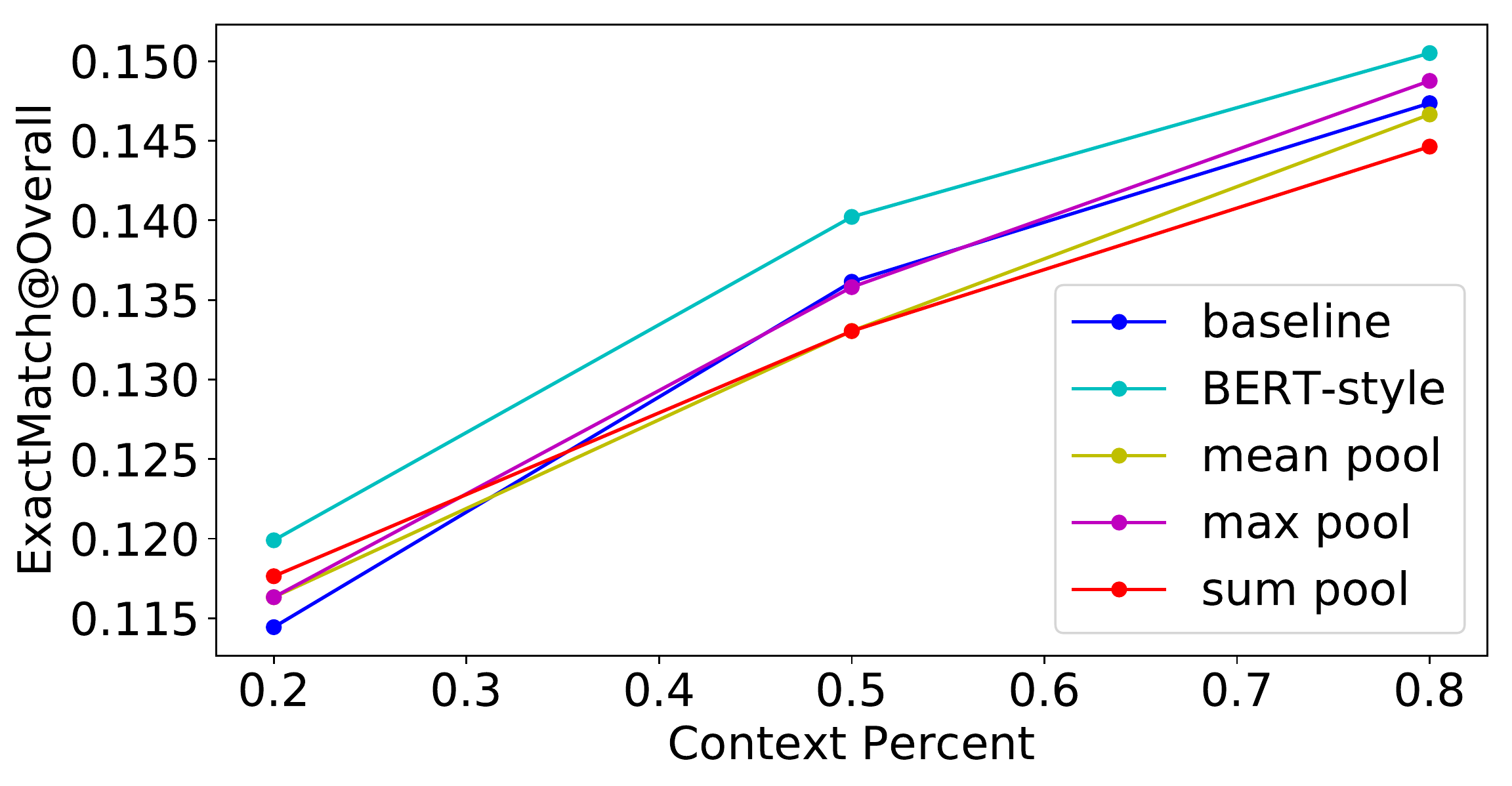}
\vspace{-0.5em}
\caption{Improvements of char. models of size $80$M with BERT-style word segment and char. pooling over baseline char. model on WikiText-103 validation set.}
\label{fig:res_bert_charavg_80M}
\vspace{-0.5em}
\end{figure}

We now discuss improvements on training character models by employing our novel methods over training a baseline character model from scratch.


\noindent\textbf{Improvements w.r.t context percent.} Figure~\ref{fig:res_bert_charavg_80M} shows improvements of character models of size $80$M with BERT-style word segment embedding and character pooling methods. Context percent corresponds to the percentage of initial tokens taken from a Wikipedia paragraph to construct the prompt, while the rest of the tokens form the ground truth. BERT-style word segment outperforms the baseline and character pooling methods on all context percent values. We attribute the inferior performance of the character pooling methods to their inability to track the order of the characters while computing the word representation. Among different pooling functions, the $max$ function performs well on most context percent values. When the context percent is very low (e.g., $0.2${-$0.35$}), it is interesting to see that all methods perform similar or outperform the baseline. This result shows that integrating word information explicitly is especially crucial when the prompts are ambiguous or contain few tokens (i.e., context percent is low). We omit the character pooling method from our further analysis due to its inferior performance. 


\begin{table}[htb]
\tiny
\begin{center}
\begin{tabular}{p{0.41in}|p{0.48in}|p{0.49in}|p{0.4in}|p{0.4in}} \toprule 
\textbf{Models}  & \textbf{Exact Match Overall (\%)} & \textbf{Partial Match Overall (\%)} & \textbf{Naturalness  (\%)} & {\textbf{Acceptability}  (\%)}  \\ \midrule 
{Human} & {100} & {100} & {88} & {100} \\ \midrule 
Base (Word) & 8.51 & 13.76 & 53 & {87} \\ \midrule 
Base (Char)  & 10.71 (+25.9\%) & 15.37 (+11.7\%) & 62 (+16.9\%) & {93 (+6.9\%)} \\ \midrule 
BERT-st. (Char) & 10.78 (+26.7\%) & 15.42 (+12.1\%) & 59 (+11.3\%) & {93 (+6.9\%)} \\ \midrule 
Transfer fr. word (Char)  & \textbf{10.83 (+27.3\%)} & \textbf{15.5 (+12.6\%)} & \textbf{69 (+30\%)} & {\textbf{94 (+8.1\%)}}   \\\bottomrule 
\end{tabular}
\vspace{-0.5em}
\caption{{Improvements of various proposed models over baseline word model of the same size ($10$M parameters) on the WikiText-103 test set.}}
\vspace{-0.5em}
\label{tab:final_res}
\end{center}
\end{table}

\noindent\textbf{{Quantitative Analysis.}}
{Table~\ref{tab:final_res} shows the performance improvements of} 
{proposed baseline character model as well as its proposed variants over baseline word model of size $10$M.} To transfer decoder layers from the word model, we first train a $20$-layer word model that has the same Transformer shape (i.e., number of heads, head dimension, model dimension, and inner dimension in feedforward layer) as the baseline word model and transfer the bottom $10\%$ of the decoder layers from the word model to initialize our character model.\footnote{The hyperparameter space for the transfer from word models method can be seen in Appendix~\ref{sec:hypspace_transfer}.} 
Consistent with the findings of \newcite{trajanovski-etal-2021-text}, we observe the improvements in ExactMatch@Overall and PartialMatch@Overall metrics to be highly correlated. {Both ``BERT-style word segment" and ``transfer from word model" methods improve upon the baseline word model by at least 26\% and 12\% (shown in Table~\ref{tab:final_res}),} 
{in terms of ExactMatch and PartialMatch respectively.}
{These methods also improve upon the baseline character model by at least 0.7\% and 0.3\% (not explicitly shown in Table~\ref{tab:final_res}),}  
{in terms of ExactMatch and PartialMatch respectively.}
Importantly, compared to the ``BERT-style word segment'' method that introduces $384$K additional parameters, our ``transfer from word model'' method does not introduce any additional parameters. This demonstrates the advantage of ``transfer from word models'' in improving baseline character model (as compared to our other methods), while leaving no impact on memory. We also perform human evaluation of suggestions generated by various autocomplete models based on their naturalness and { acceptability}. {Naturalness measures how natural the suggestion is with respect to the prompt while acceptability measures how likely the suggestion will be accepted by user (details in~\ref{sec:human_annot}).} Human suggestions taken from WikiText-103 have a naturalness {and user acceptability score} of 88\% {and 100\%} as rated by annotators. 
We observe that the ``transfer from word models'' method generates most natural {and user acceptable suggestions (69\%, 94\% resp.)}, which is better than the baseline character (62\%, {93\% resp.)} second only to the human baseline (88\%, {100\% resp.)}. 

\begin{table}[htb]
\tiny
\begin{center}
\begin{tabular}{p{2.8in}} \hline
\textbf{Prompt and Suggestions} \\ \hline 
\begin{tabular}[c]{@{}p{2.8in}}{\textbf{Prompt}}: {The Olmec civilization developed in the lowlands of southeastern Mexico ... , the Indus Valley Civilization of south Asia} \\ {\textbf{Ground truth}}: {, the civilization}\\{\textbf{Baseline}}:  {, and the} \\{\textbf{BERT-style}}: {, the indus} \\ {\textbf{Transfer from word models}}: {, the civilization} \end{tabular} \\ \hline
\begin{tabular}[c]{@{}p{2.8in}}{\textbf{Prompt}}: {Typhoon Lupit formed on November 18 from the monsoon trough to the west of the Marshall Islands . Early in its duration , it moved generally to} \\ {\textbf{Ground truth}}: {the west or}\\{\textbf{Baseline}}:  {the north of} \\{\textbf{BERT-style}}: {the west of} \\ {\textbf{Transfer from word models}}: {the west of} \end{tabular} \\ \hline
\end{tabular}
\vspace{-0.5em}
\caption{Sample suggestions of length $3$ words generated by baseline and proposed character autocomplete models. See Appendix~\ref{sec:qual_word_char} for more examples.}
\vspace{-0.5em}
\label{tab:qual_word_char_paper-1}
\end{center}
\end{table}

\noindent\textbf{Qualitative Analysis.} 
Tables~\ref{tab:qual_word_char_paper-1} and ~\ref{tab:qual_word_char} (Appendix~\ref{sec:qual_word_char}) show sample suggestions generated by the proposed baseline character autocomplete model as well as its proposed variants. 
{Suggestions generated by the strongest method seem to have better match with the ground truth and factually (e.g., direction of typhoon) correct.}\footnote{We provide a qualitative analysis of the baseline and proposed character models in the Appendix~\ref{sec:qual_analysis}.}

\section{Conclusion}
\label{sec:conclusion}
In this work, we investigated the challenging task of building autocomplete models for answering broad prompts under memory-constrained settings. To this end, we introduced some novel methods that integrate word information into a character model with negligible impact on memory. Employing our methods, we demonstrated that character models can achieve a better accuracy-memory trade-off as compared to word models.


\section{Limitations}
\label{sec:limitation}

The limitations of this work are as follows:
\begin{itemize}
    \item \textbf{English.} Our work builds autocomplete models for English language only.
    \item \textbf{Accuracy-memory tradeoff only.} Our work primarily focuses on deploying models on lower-end edge platforms where memory, as opposed to latency, is the major bottleneck. Hence, our methods may not improve the accuracy-latency tradeoff, which is a focus for future work.
    \item {\textbf{WikiText-103 dataset} Our work explores only WikiText-103 dataset for creating broad prompts. In the future, we will study other datasets (e.g., 1 Billion Word Language Model benchmark~\cite{onebword}) that explore the full range of low-frequency prompt patterns, which can arise in real-world situations.}
    \item {\textbf{Transformer-XL architecture} Our work studies only Transformer-XL architecture to build word based and character based autocomplete models. In the future, we will study other popular architectures (e.g., GPT-2~\cite{gpt_radford18}) to see the generalizability of proposed techniques.}
\end{itemize}

\section*{Acknowledgements}
\label{sec:acknow}
{MAM acknowledges support from Canada Research Chairs (CRC), the Natural Sciences and Engineering Research Council of Canada (NSERC; RGPIN-2018-04267), Canadian Foundation for Innovation (CFI; 37771), and Digital Research Alliance of Canada.\footnote{\href{https://alliancecan.ca}{https://alliancecan.ca}} Lakshmanan's research was supported in part by a grant from NSERC (Canada).}

\bibliography{anthology,custom}
\bibliographystyle{acl_natbib}

\pagebreak

\appendix
\section{Appendices}
\label{sec:appendix}

\subsection{Reproducibility}
\label{sec:reproduce}
{We experiment with both Reddit and WikiText-103 datasets. WikiText-103 is a public dataset and widely adopted as a language modeling benchmark. WikiText-103 is downloaded from \url{tinyurl.com/yajy5wjm}. The Reddit dataset used in this work is a sample of publicly available \textit{Pushshift Reddit dataset}~\cite{reddit_pushshift}. The sample contains 4M train, 20K validation and 20K test posts. The key feature of the Reddit dataset is the significantly low percentage of unique out of vocabulary n-grams compared to WikiText-103, as shown in Table~\ref{tab:oov} and discussed in Section~\ref{sec:charvsword}. For reproducibility, datasets and code used in this work is available at \url{tinyurl.com/bdd69r34} (anonymized) and will be made publicly available should paper be accepted.} 

\subsection{Hyperparameter space for computing component-wise parameter breakdown}
\label{sec:hypspace_compo}
Table~\ref{tab:hypspace_compo} displays the Transformer-XL hyperparameter space used to create 100 random architectures for computing component-wise parameter breakdown plot (Figure~\ref{fig:char-transxl-param-distrib}) for both word and character models. Rest of the hyperparameters come from the default configuration of Transformer-XL model.

\begin{table*}[htb]
\footnotesize
\begin{center}
\begin{tabular}{l|l} \hline
Hyperparameter Name & Hyperparameter Values for Sampling \\ \hline 
Number of hidden layers  & \{ 2, 4, 8, 12, 16, 24, 32 \} \\
Number of attention heads & \{ 2, 4, 8, 16, 32, 64 \} \\
Dimension of attention head & \{ 8, 16, 32, 64, 128 \} \\
Dimension of input/output embedding & \{ 256, 512, 1024, 2048 \} \\
Inner dimension of feedforward layer & \{ 256, 512, 1024, 2048 \} \\
Dimension of model & \{ 256, 512, 1024, 2048 \} \\ \hline
\end{tabular}
\caption{Hyperparameter space for computing component-wise parameter breakdown for both word and character models.}
\label{tab:hypspace_compo}
\end{center}
\end{table*}

\subsection{Hyperparameter values for word and character models of different sizes}
\label{sec:hypspace_word_char}
Table~\ref{tab:hypspace_word} displays the hyperparameter values for word models of different sizes used in the paper. Table~\ref{tab:hypspace_char} displays the hyperparameter values for character models of different sizes used in the paper. 

\begin{table*}[htb]
\footnotesize
\begin{center}
\begin{tabular}{p{2.5in}|l|l|l|l|l|l|l} \hline
Hyperparameter name / Model size & $5M$ & $10M$ & $20M$ & $30M$ & $40M$ & $50M$ & $80M$  \\ \hline 
Number of hidden layers  &  3 & 4 & 6 & 12 & 14 & 16 & 16 \\
Number of attention heads &  4 & 4 & 8 & 8 & 8 & 8  & 32 \\
Dimension of attention head &  24 & 24 & 32 & 32 & 32 & 32 & 32 \\
Dimension of input/output embedding & 18 & 36 & 74 & 100 & 128 & 160 & 256\\
Inner dimension of feedforward layer & 60 & 150 & 200 & 768 & 900 & 800 & 768 \\
Dimension of model & 18 & 36 & 74 & 100 & 128 & 160 & 256 \\ 
Number of tokens to predict during training & 192 & 192 & 192 & 192 & 192 & 192 & 192 \\
Number of tokens cached from previous iterations during training & 192 & 192 & 192 & 192 & 192 & 192 & 192 \\
Learning rate & 0.01 & 0.01 & 0.01 & 0.01 & 0.01 & 0.01 & 0.01  \\
Number of iterations for learning rate warmup & 1K & 1K & 1K & 1K & 1K & 1K & 1K \\
Maximum number of training steps & 200K & 200K & 200K & 200K & 200K & 200K & 200K  \\
Batch size & 256 & 256 & 256 & 256 & 256 & 256 & 256 \\
Number of tokens to predict during evaluation & 192 & 192 & 192 & 192 & 192 & 192 & 192  \\
Number of tokens cached from previous iterations during evaluation & 192 & 192 & 192 & 192 & 192 & 192 & 192  \\ 
Vocabulary size & 267736 & 267736 & 267736 & 267736 & 267736 & 267736 & 267736 \\ \hline
\end{tabular}
\caption{Hyperparameter values for word models of different sizes.}
\label{tab:hypspace_word}
\end{center}
\end{table*}

\begin{table*}[htb]
\footnotesize
\begin{center}
\begin{tabular}{p{3.5in}|l|l|l|l} \hline
Hyperparameter name / Model size & $5M$ & $10M$ & $20M$ & $80M$ \\ \hline 
Number of hidden layers  & 12 & 12 & 12 & 16  \\
Number of attention heads & 8 & 8 & 8 & 8 \\
Dimension of attention head & 32 & 32 & 64 & 64 \\
Dimension of input/output embedding & 278 & 512 & 550 & 750 \\
Inner dimension of feedforward layer & 128 & 165 & 250 & 2048 \\
Dimension of model &  278 & 512 & 550 & 750  \\ 
Number of tokens to predict during training & 512  & 512  & 512  & 512 \\
Number of tokens cached from previous iterations during training & 512  & 512  & 512  & 512  \\
Learning rate & 0.001 & 0.001  & 0.001  & 0.001     \\
Number of iterations for learning rate warmup & 4K & 4K & 4K & 4K  \\
Maximum number of training steps & 400K & 400K & 400K & 400K   \\
Batch size & 128 & 128  & 128  & 128  \\
Number of tokens to predict during evaluation & 512  & 512  & 512  & 512  \\ 
Number of tokens cached from previous iterations during evaluation & 2K & 2K & 2K & 2K \\
Vocabulary size & 128 & 128 & 128 & 128 \\ \hline
\end{tabular}
\caption{Hyperparameter values for character models of different sizes.}
\label{tab:hypspace_char}
\end{center}
\end{table*}

\subsection{Hyperparameter space for transfer from word models method}
\label{sec:hypspace_transfer}
Table~\ref{tab:hypspace_compo} displays the hyperparameter space for the proposed transfer from word models method.

\begin{table*}[htb]
\footnotesize
\begin{center}
\begin{tabular}{l|l} \hline
Hyperparameter Name & Hyperparameter Values \\ \hline 
Number of hidden layers  & \{ 4, 8, 12, 16, 20, 24 \} \\
Percentage of bottom most layers to transfer  & \{ 10\%, 20\%, 30\%, 40\%, 50\% \} \\ \hline
\end{tabular}
\caption{Hyperparameter space for transfer from word models method.}
\label{tab:hypspace_compo}
\end{center}
\end{table*}

\subsection{Greedy vs. Beam search decoding}
\label{sec:beam_search_analysis}
\begin{figure*}[t!]
    \centering
    \begin{subfigure}[t]{0.4\textwidth}
        \centering
\includegraphics[height=1.25in, width=2.0in]{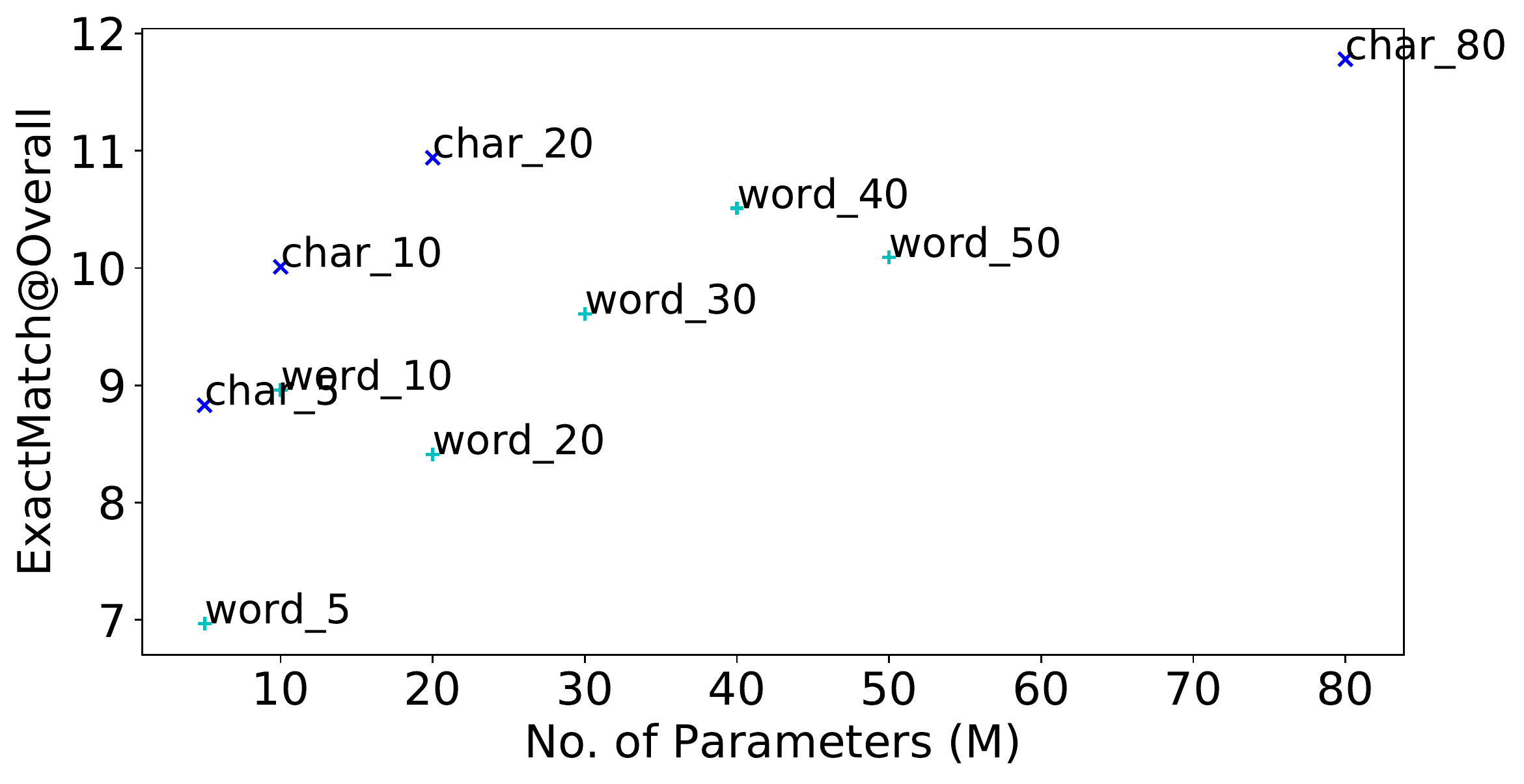}
        \caption{Greedy search}
    \end{subfigure}%
    ~ 
    \begin{subfigure}[t]{0.4\textwidth}
        \centering
        \includegraphics[height=1.25in, width=2.0in]{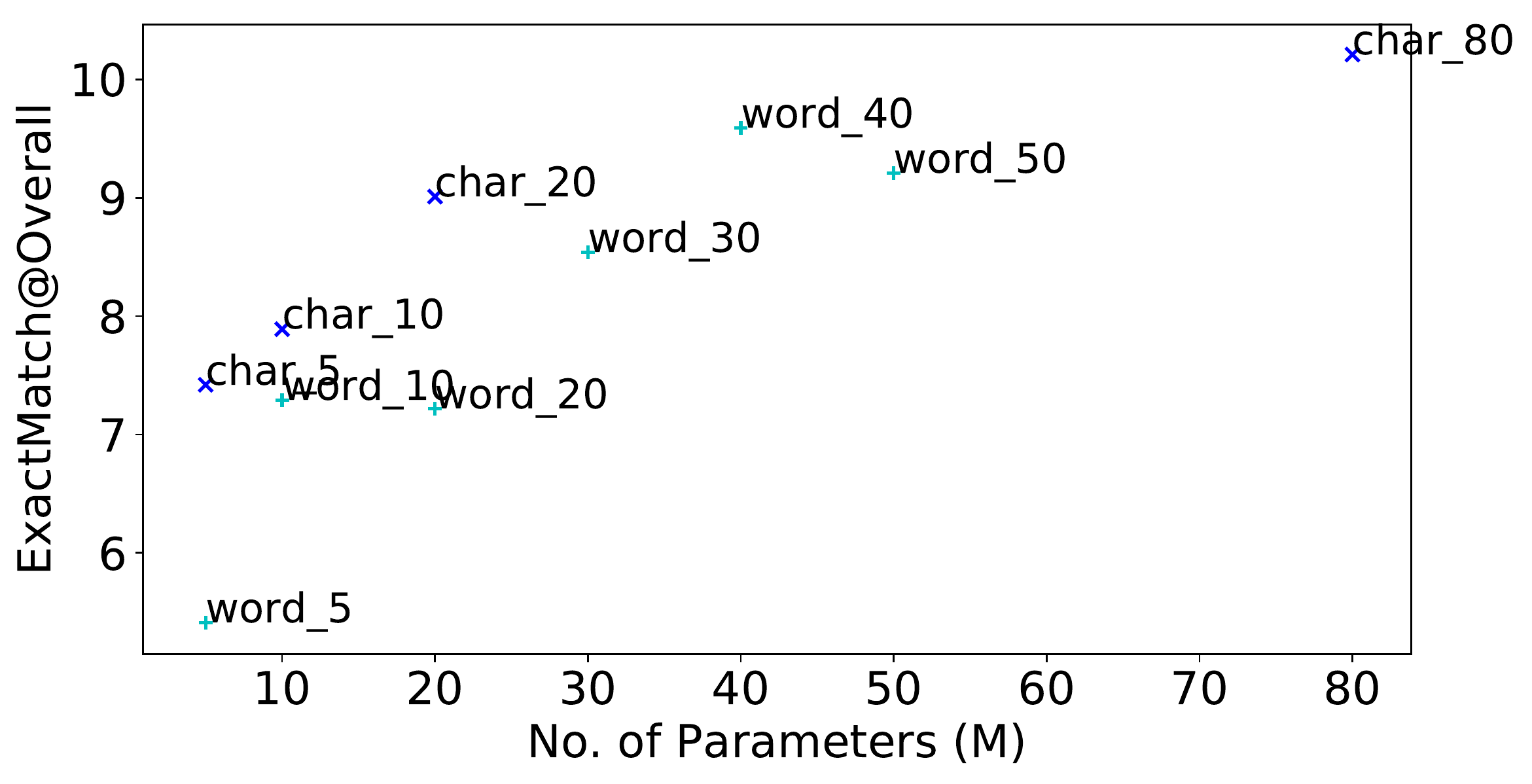}
        \caption{Beam search}
    \end{subfigure}
    \caption{Greedy search vs. Beam search on WikiText-103 test set. Beam size and prompt context percentage is set as 5 and 20\% respectively.}
    \label{fig:beam_search_analysis}
\end{figure*}

Figure~\ref{fig:beam_search_analysis} shows the pareto-curve for greedy and beam search. It is clear that smaller character models rival bigger word models regardless of the choice of decoding algorithm. Strikingly, we find greedy search to outperform beam search by a large margin. Two possible reasons are: (i) the noise injected by the adaptive softmax approximation of predicted probability distribution over vocabulary, and/or (ii) sensitivity of beam search to explore spurious hypothesis when the user prompt patterns are low frequency.

\begin{figure}
\centering
\includegraphics[width=2.3in,height=1.25in]{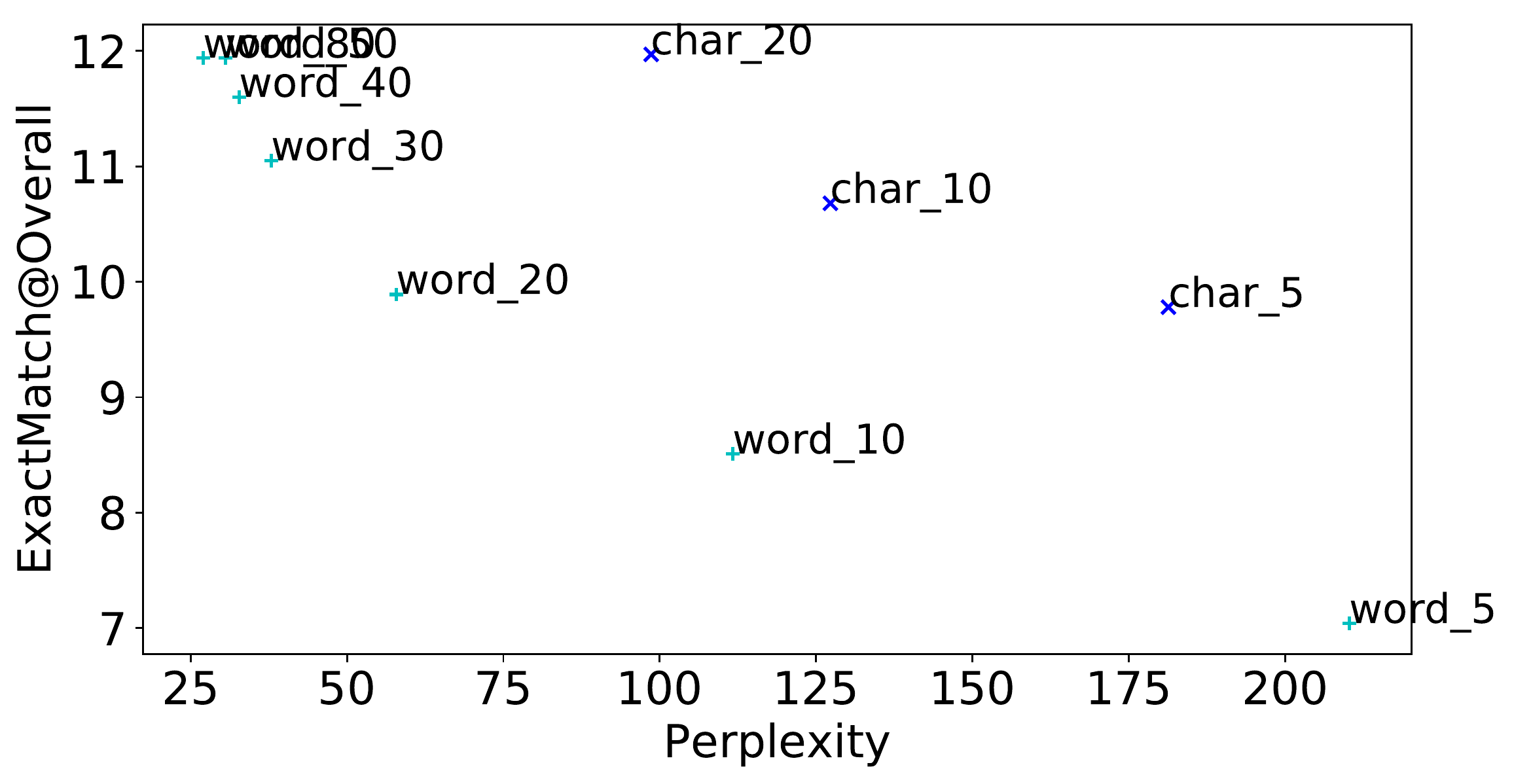}
\caption{Perplexity vs. ExactMatch. For comparison, perplexity output by character models (also known as bits per byte) is converted to perplexity per word using the formula proposed in \newcite{Choe2019BridgingTG}.}
\label{fig:ppl_vs_em_diff}
\end{figure}

\subsection{Differences of Autocomplete from Conventional Language Modeling Task.}
\label{sec:ppl_vs_em}
{The  autocomplete task is a well-defined problem with rich prior literature (see Section~\ref{sec:related_work}). Existing autocomplete research, including ours, is focused on building a conventional language model that computes the likelihood of a text sequence.} The training procedure for our autocomplete task and that for conventional language modeling (CLM) task are generally similar. However, 
the goal of our autocomplete task is to generate suggestions with high precision (as captured by ExactMatch) while the main goal of CLM is to maximize the overall data likelihood (as captured by perplexity).~\newcite{smartcompose_kdd19} show that perplexity and ExactMatch metrics are only weakly correlated as improvements in perplexity could be ``mostly in places where the model is relatively low in {likelihood} score''. As shown in Figure~\ref{fig:ppl_vs_em_diff}, autocomplete models with poorer perplexity scores (e.g., character model of size $20$M) can enjoy better ExactMatch scores compared to models with better perplexity scores (e.g., word model of size $20$M). We also perform a theoretical analysis to show how perplexity scores can change drastically for the same ExactMatch score (details in Appendix~\ref{sec:ppl_vs_em_theory}). {Thus, building a good language model is not enough to solve the autocomplete task.} 
{Another major conceptual difference between CLM and autocomplete tasks is that the former focuses mainly on generating long horizon (typically 128-512 tokens) continuation while the latter focuses on generating short horizon (typically 3-5 tokens) continuation.}

\subsection{Theoretical analysis on differences in perplexity and Exact Match metrics}
\label{sec:ppl_vs_em_theory}
We will conduct a theoretical study to show the differences in the information captured by perplexity and Exact Match metric. Specifically, we show that the exact match score can be perfect whereas perplexity score can  either be perfect or worsen  by a large margin (\textbf{Claim 1}). Conversely, we also show that the exact match score can be the worst (i.e., zero) whereas the perplexity score can be  poor or better by a large margin (\textbf{Claim 2}). Without loss of generality, we assume the vocabulary size $\mathcal{V}$ to be 2. Let $A$, $B$ be the two tokens corresponding to the first and second index in the vocabulary respectively. Consider a single token prediction ($\hat{x}_j$) and let the ground truth token be $B$, that is, $\hat{x}_j = [0, 1]$. Table~\ref{tab:ppl_vs_em_theory} shows the differences in perplexity score and Exact Match score as a  function of $\hat{x}_j$, as it varies slightly. The first six rows in the table validate \textbf{Claim 1}, where exact match score is 1 but the perplexity ranges $-9.9\mathrm{e}{-10}$ to $0.67$. The rest of the rows validate \textbf{Claim 2}, where the exact match score is 0 but the perplexity score ranges from $0.69$  to $20.72$.

\begin{table*}[htb]
\footnotesize
\begin{center}
\begin{tabular}{cccc} \toprule
Ground truth ($x_j$) & Prediction ($\hat{x}_j$) & Exact Match & Perplexity \\ \midrule
$[0, 1]$  & $[0, 1]$ & 1 & $-9.9\mathrm{e}{-10}$ \\
$[0, 1]$  & $[0.1, 0.9]$ & 1 & 0.11 \\
$[0, 1]$  & $[0.2, 0.8]$ & 1 & 0.22 \\
$[0, 1]$  & $[0.3, 0.7]$ & 1 & 0.36 \\
$[0, 1]$  & $[0.4, 0.6]$ & 1 & 0.51 \\
$[0, 1]$  & $[0.49, 0.51]$ & 1 & 0.67\\
$[0, 1]$  & $[0.5, 0.5]$ & 0 & 0.69 \\
$[0, 1]$  & $[0.51, 0.49]$ & 0 & 0.71\\
$[0, 1]$  & $[0.6, 0.4]$ & 0 & 0.92 \\
$[0, 1]$  & $[0.7, 0.3]$ & 0 & 1.2 \\
$[0, 1]$  & $[0.8, 0.2]$ & 0 & 1.61 \\
$[0, 1]$  & $[0.9, 0.1]$ & 0 & 2.3 \\
$[0, 1]$  & $[1.0, 0]$ & 0 & 20.72 \\ \bottomrule
\end{tabular}
\caption{Differences in perplexity and Exact Match as function of small changes in $\hat{x}_j$ when the ground truth is $[0, 1]$.}
\label{tab:ppl_vs_em_theory}
\end{center}
\end{table*}

\subsection{Accuracy-Memory Pareto-Curve on Unseen Datasets}
\label{sec:acc-vs-mem-unseen-datasets}
\begin{figure*}[t!]
    \centering
    \begin{subfigure}[t]{0.4\textwidth}
        \centering
\includegraphics[height=1.25in, width=2.0in]{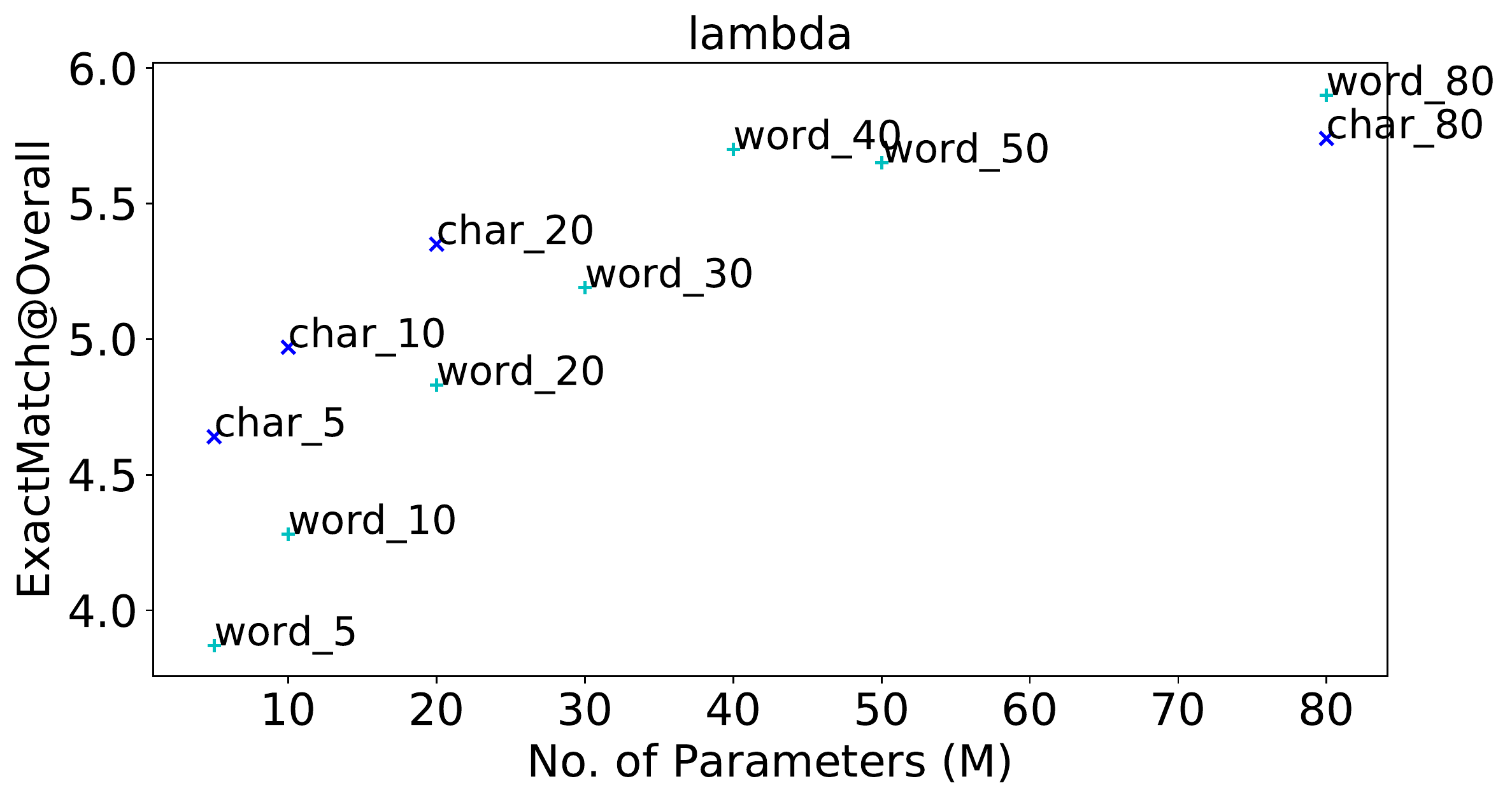}
        \caption{LAMBADA~\cite{paperno-etal-2016-lambada}}
    \end{subfigure}%
    ~ 
    \begin{subfigure}[t]{0.4\textwidth}
        \centering
        \includegraphics[height=1.25in, width=2.0in]{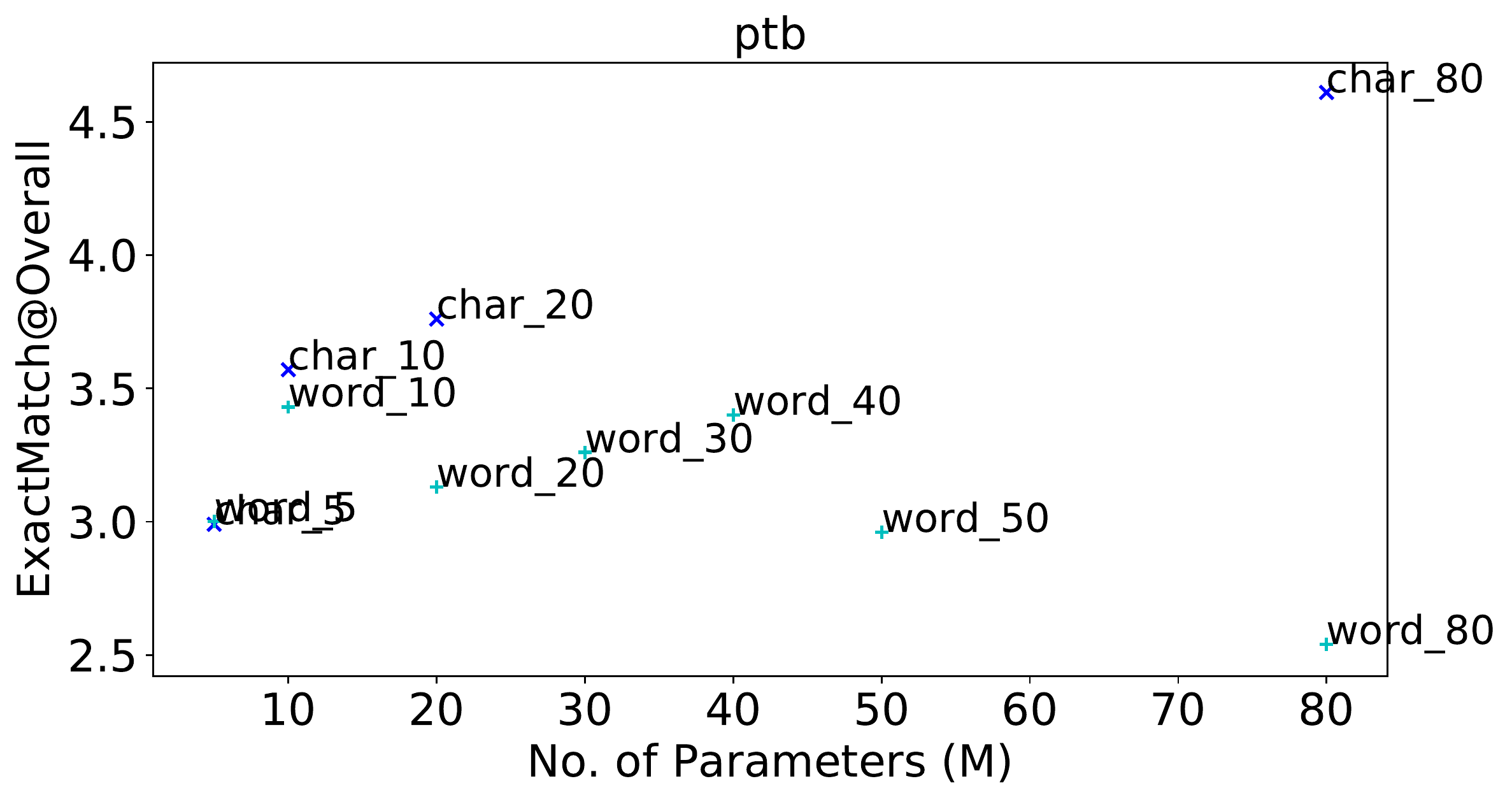}
        \caption{PTB~\cite{ptb}}
    \end{subfigure}
    \caption{Accuracy-Memory Pareto Curve for Autocomplete models trained on WikiText-103 and evaluated on test set of two unseen datasets: LAMBADA and PTB.}
    \label{fig:acc-vs-mem-unseen-datasets}
\end{figure*}

We study the accuracy-memory pareto curve of autocomplete models trained on WikiText-103 and evaluate on the test set of two unseen datasets: LAnguage Modeling Broadened to Account for Discourse Aspects~\cite{paperno-etal-2016-lambada} (LAMBADA, mostly focused prompts) and Penn Treebank~\cite{ptb} (PTB, mostly broad prompts). From  Figure~\ref{fig:acc-vs-mem-unseen-datasets}, we observe that the trend where smaller character models rival larger word models that holds true for answering broad prompts (PTB) but not clearly for answering focused prompts (LAMBADA). It is striking that the trend holds true for broad prompts even when the examples are unseen during the training of the autocomplete model.

\subsection{Qualitative examples of suggestions from autocomplete models}
\label{sec:qual_word_char}
Table~\ref{tab:qual_word_char} displays sample suggestions generated by vanilla and proposed character autocomplete models, grouped by the type of artifact in the generation.

\begin{table*}[htb]
\footnotesize
\begin{center}
\begin{tabular}{p{0.58in}p{5.5in}} \hline
\textbf{Artifact type} & \textbf{Prompt and Suggestions} \\ \hline 
Plausible & \begin{tabular}[c]{@{}p{5.5in}@{}}{\textbf{Prompt}}: In 2006 Boulter starred in the play Citizenship written by Mark Ravenhill . The play was part of a series which featured different playwrights , titled Burn / Chatroom / Citizenship . In a 2006\\ {\textbf{Ground truth}}: interview , fellow\\{\textbf{Baseline}}: interview , ravenhill \\ {\textbf{BERT-style}}: interview with the \\ {\textbf{Transfer from word models}}: interview with the \end{tabular} \\ \hline
Plausible & \begin{tabular}[c]{@{}p{5.5in}@{}}{\textbf{Prompt}}: In December 759 , he briefly stayed in Tonggu ( modern Gansu ) . He departed on December 24 for Chengdu ( Sichuan province ) , where he was hosted by local Prefect and \\ {\textbf{Ground truth}}: fellow poet Pei\\{\textbf{Baseline}}: servant and served \\ {\textbf{BERT-style}}: chief executive officer \\ {\textbf{Transfer from word models}}: commissioned as a \end{tabular} \\ \hline
Semantic error & \begin{tabular}[c]{@{}p{5.5in}@{}}{\textbf{Prompt}}: In his lifetime and immediately following his death , Du Fu was not greatly appreciated . In part this can be attributed to his stylistic and formal innovations, some of which are still "considered extremely daring and bizarre by Chinese critics ." There are few contemporary references to him — only eleven poems from six writers — and these describe him in terms of affection, but not as a \\ {\textbf{Ground truth}}: paragon of poetic\\{\textbf{Baseline}}: reference to his  \\ {\textbf{BERT-style}}: {\color{red}{poem}} . the \\ {\textbf{Transfer from word models}}: consequence of his \end{tabular} \\ \hline
Semantic error & \begin{tabular}[c]{@{}p{5.5in}@{}}{\textbf{Prompt}}: Other translators have placed much greater weight on trying to convey a sense of the poetic forms used by Du Fu . Vikram Seth in Three Chinese Poets uses English @-@ style rhyme schemes , whereas Keith Holyoak in Facing the Moon approximates the Chinese rhyme scheme ; both use end @-@ stopped lines and preserve some degree of parallelism . In The Selected Poems of Du Fu , Burton Watson follows the parallelisms quite strictly , persuading the western reader to adapt to the poems rather than \\ {\textbf{Ground truth}}: vice versa .\\{\textbf{Baseline}}: {\color{red}{to the poems}}  \\ {\textbf{BERT-style}}: {\color{red}{adapt the poems}} \\ {\textbf{Transfer from word models}}: {\color{green}{the parallelisms}} of \end{tabular} \\ \hline
Repetition & \begin{tabular}[c]{@{}p{5.5in}@{}}{\textbf{Prompt}}: Although initially he was little @-@ known to other writers , his works came to be hugely influential in both \\ {\textbf{Ground truth}}: Chinese and Japanese\\{\textbf{Baseline}}: the writers and  \\ {\textbf{BERT-style}}: {\color{red}{writers and writers}} \\ {\textbf{Transfer from word models}}: the ancient and \end{tabular} \\ \hline
Repetition & \begin{tabular}[c]{@{}p{5.5in}@{}}{\textbf{Prompt}}: In the 20th century , he was the favourite poet of Kenneth \\ {\textbf{Ground truth}}: Rexroth , who \\{\textbf{Baseline}}: {\color{red}{kenneth kenneth kenneth}}  \\ {\textbf{BERT-style}}: county . the \\ {\textbf{Transfer from word models}}: {\color{red}{kenneth kenneth kenneth}}  \end{tabular} \\ \hline
Grammatical error & \begin{tabular}[c]{@{}p{5.5in}@{}}{\textbf{Prompt}}: Hung summarises his life by concluding that , \\ {\textbf{Ground truth}}: " He appeared \\{\textbf{Baseline}}: according to {\color{red}{ksummarises}}  \\ {\textbf{BERT-style}}: in the same \\ {\textbf{Transfer from word models}}: as a result  \end{tabular} \\ \hline
\end{tabular}
\caption{Sample suggestions of length $3$ words generated by vanilla and proposed character autocomplete models, grouped by the type of artifact in the generation.}
\label{tab:qual_word_char}
\end{center}
\end{table*}

\subsection{Qualitative analysis of vanilla and proposed character models}
\label{sec:qual_analysis}

\begin{table*}[]
\footnotesize
\begin{center}
\begin{tabular}{p{1.0in}p{0.5in}p{1.0in}p{1.5in}} \toprule
\textbf{Artifact type} & \textbf{Baseline} & \textbf{BERT-style w. seg.} & \textbf{Transfer from word models} \\ \midrule
Plausible ($\uparrow$) & 40 & 40 & \textbf{42}  \\ 
Semantic Error ($\downarrow$) & 7 & \textbf{6} & 7 \\
Repetition ($\downarrow$) & 7 & 7 & \textbf{5} \\
Gram. Error ($\downarrow$) & 3 & 3 & \textbf{2} \\ \bottomrule
\end{tabular}
\caption{Percentage of different artifacts in the generated suggestion from vanilla and proposed character models, by manual inspection of $100$ WikiText-103 examples. $\uparrow$ indicates higher the better, $\downarrow$ indicates lower the better.}
\label{tab:qual_res}
\end{center}
\end{table*}
We manually inspect the suggestions generated by vanilla and proposed character models\footnote{Sample suggestions from different autocomplete models can be seen in Appendix~\ref{sec:qual_word_char}.}.  Table~\ref{tab:qual_res} displays the percentage of different artifacts: \textit{plausible} (plausible suggestion that does not have exact match with the ground truth), \textit{semantic error} (e.g., new n-gram, incorrect n-gram usage), \textit{repetition} (e.g., n-gram with repetitions), and \textit{grammatical error}. Compared to baseline and BERT-style word segment model, character model with decoder layer transfer from word model results in less undesirable artifacts overall. 

\subsection{Human annotation of suggestions}
\label{sec:human_annot}
We conduct human annotation of suggestions outputted by various autocomplete models based on \textit{naturalness} (how natural the suggestion is with respect to the prompt?) {and \textit{acceptability} (whether the suggestion will be accepted by user or not?).} Some aspects of natural suggestion are borrowed from \newcite{dou-etal-2022-gpt}. The annotation guideline for naturalness and {acceptability} can be seen in Table~\ref{tab:annotguideline} and {Table~\ref{tab:annotguidelineacceptability} respectively.} We ask 8 annotators to rate 10 suggestions each.

\begin{table*}[htb]
\footnotesize
\begin{center}
\begin{tabular}{|p{5in}|} \hline
Autocomplete is a task where the user inputs a text, which is conditioned by the model to generate `natural' continuation (or suggestion). The goal of this annotation effort is to rate the quality of suggestions generated by various autocomplete models based on the `natural'ness. Each suggestion will be at most three words. Keep in mind that there could be more than one `natural' suggestion for a text. \\ \\

Some aspects of suggestion (but don't restrict only to these) that makes a suggestion NOT natural can be: grammatical error (missing words, extra words, incorrect or out of order words), redundancy (extra unnecessary information, word repetition), off-prompt (suggestion is unrelated to the text), self-contradiction (suggestion contradicts the text), incoherence (grammatical, not redundant, on prompt, not contradictory but still CONFUSING), factual or commonsense errors (violates our basic understanding of the world) and so on. Assume a broad definition of `natural'ness and use your best judgement to rate. \\ \\ 

You will be asked to annotate TEN texts. For each text, you will see a suggestion and you will rate by picking exactly one of the two choices: \\
(i) natural - Select this option if suggestion is natural with respect to the text \\
(ii) NOT natural - Select this option if suggestion is NOT natural with respect to the text \\ \hline
\end{tabular}
\caption{Annotation guideline for human annotators to rate the quality of suggestions generated by autocomplete models and humans based on naturalness.}
\label{tab:annotguideline}
\end{center}
\end{table*}

\begin{table*}[htb]
\footnotesize
\begin{center}
\begin{tabular}{|p{5in}|} \hline
Autocomplete is a task where a user inputs a text (prompt), which is conditioned by the model to generate `natural' continuation (or suggestion). For example, the user can give the prompt ``Filmmaker George Lucas used Tikal as a'', and the system may give a suggestion such as ``filming location''. An autocomplete system is successful if it can reduce the keystrokes a user would need to make, improving user productivity. The goal of this annotation task is to decide if (i) a suggestion generated by an autocomplete model will be accepted by a user (to reduce the keystrokes) or (ii) not. Each suggestion will be at most three words. \\ \\
 
You can accept the suggestion if it is useful. A suggestion can be useful for one or more reasons (but don't restrict only to these): (i) the suggestion seems completely relevant to the prompt; (ii) the suggestion can be minimally edited for it to be useful. Note that reasons for acceptability are generally subjective. Hence, please assume a broad definition of “usefulness” and employ your best judgment to rate. \\ \\
 
You will be asked to annotate 10 texts. For each text, you will see a suggestion and you will rate by picking exactly one of the two choices: \\
(i) yes - Select this option if you will accept the suggestion \\
(ii) no - Select this option if you will not accept the suggestion \\ \\

The following is an example: \\

Filmmaker George Lucas used Tikal as a \\
\textbf{Suggestion:} filming location \\
\textbf{Rating choices:} \\
(i) yes - Select this option if you will accept the suggestion \\
(ii) no - Select this option if you will not accept the suggestion \\
Rating [type 'yes' or 'no' here in this line]: yes \\ \hline

\end{tabular}
\caption{{Annotation guideline for human annotators to rate the quality of suggestions generated by autocomplete models and humans based on acceptability.}}
\label{tab:annotguidelineacceptability}
\end{center}
\end{table*}

\end{document}